\documentclass[letterpaper, 10 pt, conference]{ieeeconf}  
\usepackage{graphicx}
\usepackage{subfigure}
\usepackage{multirow}
\usepackage{threeparttable}
\graphicspath{ {./pics/} }
\usepackage{cite}

\usepackage{url}
\usepackage{algorithm}

\IEEEoverridecommandlockouts                              
\title{\LARGE \bf
SUPS: A Simulated Underground Parking Scenario Dataset for Autonomous Driving
}

\author{Jiawei Hou$^1$, Qi Chen$^1$, Yurong Cheng$^1$, Guang Chen$^2$, Xiangyang Xue$^3$, Taiping Zeng$^3$, Jian Pu$^{3\ast}$\\
$^1$ School of Computer Science, Fudan University, Shanghai, China.\\
$^2$ FAW (Nanjing) Technology Development Co.,Ltd, Nanjing, China.\\
$^3$ Institute of Science and Technology for Brain-Inspired
Intelligence, Fudan University, Shanghai, China.
}

\begin{document}
\twocolumn[{
\renewcommand\twocolumn[1][]{#1}
\maketitle
\begin{figure}[H]
\hsize=\textwidth
\centering
\includegraphics[width=\textwidth]{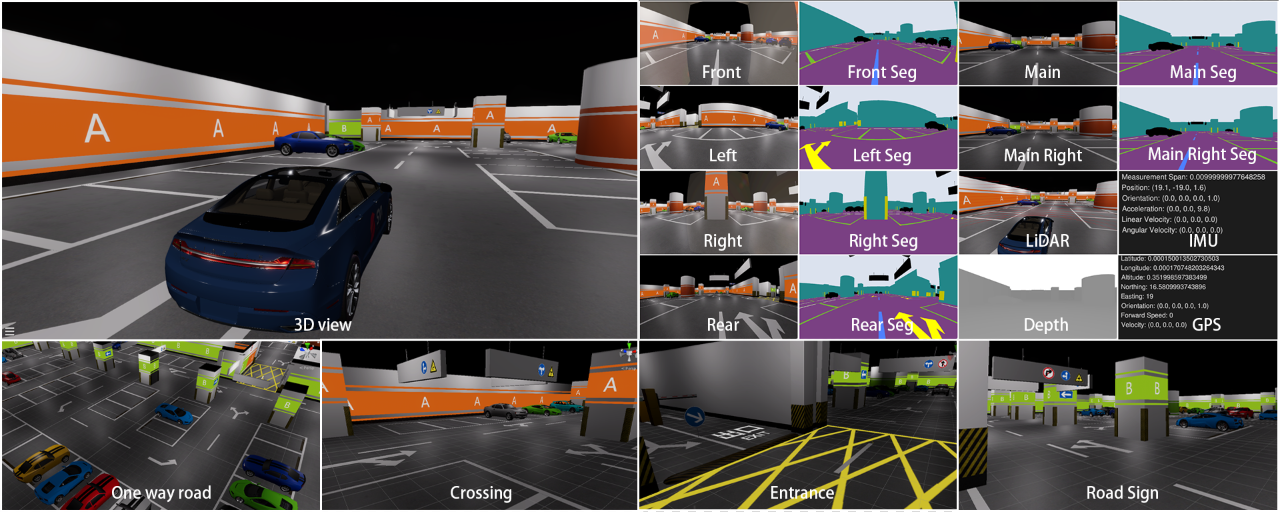}
\caption{Example of outputs from a complete set of sensors. The top left corner is a 3D view of the vehicle in the simulation scene. There are four columns of images on the top right corner, and from left to right, the first column shows frames captured by four fisheye cameras, the second column shows semantic segmentation ground truth captured by corresponding semantic cameras of the previous column, the third column shows frames from the forward stereo pinhole cameras, LiDAR scan and depth camera, and the last column shows segmentation ground truth of stereo cameras and measurements from IMU and GPS. The four pictures below show some delicate scenarios in the virtual scene, such as the one-way road, crossing, entrance, and road signs.}
\label{full_sensor}
\end{figure}
}]
\thispagestyle{empty}
\pagestyle{empty}
\begin{abstract}
Automatic underground parking has attracted considerable attention as the scope of autonomous driving expands. The auto-vehicle is supposed to obtain the environmental information, track its location, and build a reliable map of the scenario. Mainstream solutions consist of well-trained neural networks and simultaneous localization and mapping (SLAM) methods, which need numerous carefully labeled images and multiple sensor estimations. However, there is a lack of underground parking scenario datasets with multiple sensors and well-labeled images that support both SLAM tasks and perception tasks, such as semantic segmentation and parking slot detection. In this paper, we present SUPS, a simulated dataset for underground automatic parking, which supports multiple tasks with multiple sensors and multiple semantic labels aligned with successive images according to timestamps. We intend to cover the defect of existing datasets with the variability of environments and the diversity and accessibility of sensors in the virtual scene. Specifically, the dataset records frames from four surrounding fisheye cameras, two forward pinhole cameras, a depth camera, and data from LiDAR, inertial measurement unit (IMU), GNSS. Pixel-level semantic labels are provided for objects, especially ground signs such as arrows, parking lines, lanes, and speed bumps. Perception, 3D reconstruction, depth estimation, and SLAM, and other relative tasks are supported by our dataset. We also evaluate the state-of-the-art SLAM algorithms and perception models on our dataset. Finally, we open source our virtual 3D scene built based on Unity Engine and release our dataset at https://github.com/jarvishou829/SUPS.

\end{abstract}
\begin{table*}[tp]
\caption{Summary of various autonomous driving datasets with configurations and tasks}
\label{table_ex}
\begin{tabular}{llccccccc}
\hline
Task/Info & Quantity & PS2.0\cite{PS2.0} & Cityscpes\cite{Cityscapes} & TUM VI\cite{TUM} & KITTI\cite{kitti} & WoodScape\cite{woodscape} & VIODE\cite{VIODE} & SUPS(Ours) \\  
\hline
\hline
\multirow{4}[10]{*} {\begin{tabular}[l]{@{}l@{}}Capture\\ Information\end{tabular}} & Year & 2018 & 2016 & 2018 & 2012/14/15 & 2018/19 & 2021 & 2021/22 \\
                                    & Carrier & Car & Car & UAV & Car & Car & UAV* & Car \\
                                    & \begin{tabular}[l]{@{}l@{}}Other \\ sensors\end{tabular} & - & - & IMU & \begin{tabular}[c]{@{}c@{}}1 LiDAR\\ GPS\end{tabular} & \begin{tabular}[c]{@{}c@{}}1 LiDAR\\ GNSS\\ IMU\end{tabular} & \begin{tabular}[c]{@{}c@{}}IMU*\\ Segmentation\end{tabular} & \begin{tabular}[c]{@{}c@{}}1 LiDAR*\\ GPS*,IMU*\\Segmentation*\end{tabular} \\
                                    & \begin{tabular}[l]{@{}l@{}}Ground-truth\\ Trajectory \end{tabular} & - & - &\begin{tabular}[c]{@{}c@{}}MoCap\\ (partial)\end{tabular}& \begin{tabular}[c]{@{}c@{}}Fused IMU,\\ GNSS\end{tabular} & \begin{tabular}[c]{@{}c@{}}Fused IMU,\\ GNSS\end{tabular} & Simulation & Simulation \\
\hline
\multirow{2}[3]{*}{\begin{tabular}[l]{@{}l@{}}Camera\\ Information\end{tabular}} & Cameras & - & 2 & 2 & 4 & 4 & 1 & 6 \\
                                    & Type & \begin{tabular}[c]{@{}c@{}}Surround\\ view\end{tabular} & - & Stereo & Stereo & \begin{tabular}[c]{@{}c@{}}Surround\\ view\end{tabular} & Stereo* & \begin{tabular}[c]{@{}c@{}}Stereo*\\ Surround view*\end{tabular} \\
\hline \begin{tabular}[l]{@{}l@{}}Parking Slot\\ Detection\end{tabular} & Support & $\surd$ & - & - & - & - & - & $\surd$ \\
\hline \begin{tabular}[l]{@{}l@{}}Semantic\\ Segmentation\end{tabular} & Support & - & $\surd$ & - & $\surd$ & $\surd$ & $\surd$ & $\surd$ \\
\hline \begin{tabular}[l]{@{}l@{}}Depth\\ Estimation\end{tabular} & Support & - & - & $\surd$ & $\surd$ & $\surd$ & - & $\surd$  \\
\hline Visual SLAM & Support & - & - & $\surd$ & $\surd$ & $\triangle$ & $\surd$ & $\surd$ \\
\hline LOAM SLAM & Support & - & - & - & - & $\triangle$ & - & $\surd$ \\
\hline  
\end{tabular}
\begin{tablenotes}
\footnotesize
    \item []{${\rm \triangle}$To the best of our knowledge, some sequences for SLAM tasks are not open-source yet in WoodScape\cite{woodscape}}
    \item[]{$^{\rm *}$Simulated carrier and sensor.}
\end{tablenotes}
\end{table*}  
\section{INTRODUCTION}
Automatic underground parking is one of the essential parts of autonomous driving. During this process, the auto-vehicle is proposed to obtain the environmental information from the underground parking scenario, process the input from sensors, and make planning. To address this complete procedure, various algorithms are needed, such as semantic segmentation, parking slot detection, simultaneous localization and mapping (SLAM), etc. However, applying these algorithms could be challenging in the underground parking scenario. In such a scenario, narrow roads with obstacles, areas with dazzling or dim lighting, and poor-texture walls increase the noise and uncertainty of sensor measurements\cite{underpark}. The structure and content of the environment are highly repetitive compared to the urban street scene with landmarks, which makes it hard for feature matching and place recognition. In addition, vehicles may obtain an unreliable estimation result for location due to the Global Navigation Satellite System (GNSS) signal shielding, dim conditions, and ground reflection\cite{gnssa}\cite{gnssb}.

Recently, researchers have found that these challenges can be partly solved by applying deep learning to help vehicles perceive the surrounding scenario\cite{gcn} \cite{nvidia} \cite{lidar_surrounding}. Specifically, semantic information increases the perception ability of the autonomous car, and in return, an accurate pose estimation may also help perception. However, many commonly used datasets provide either successive images with timestamps for the SLAM task or discrete images for visual perception tasks, which makes it difficult for users to develop a functional automatic parking system using a single dataset. Besides, there is a large density of obstacles and ground sign semantic information in underground parking scenarios, including walls, pillars, drivable areas, lanes, parking slots, arrows, and bumps. Only a few existing datasets completely labeled these items, which requires plenty of human labor and time.

To cover the lack of datasets that support both the SLAM and perception tasks with well-labeled data, in this paper, we present SUPS, a novel dataset supporting multiple tasks through multiple sensors and multiple semantic labels to address the issues in developing an underground automatic parking system. The virtual scene used in SUPS is simulated through LGSVL \cite{lgsvl}, which is an autonomous vehicle simulation platform with highly flexible configurations of vehicles and sensors. The simulation platform has the advantage of environmental variability as well as the virtual sensor diversity and accessibility. The virtual underground parking scenario we built is similar to the real-world underground parking scenario in illumination, texture, content, and complexity of scenarios. Multiple sensors such as fisheye cameras, pinhole cameras, depth cameras, LiDARs, inertial measurement unit(IMU), GNSS are activated among records. Pixel-level semantic segmentation, parking slot detection, depth estimation, visual SLAM, and LiDAR-based SLAM tasks are supported in this dataset.

Our main contributions of this work are summarized as follows:
\begin{itemize}
    \item We present SUPS, a simulated dataset for underground automatic parking, which supports multiple tasks with multiple sensors and multiple semantic labels aligned with successive images according to timestamps. It allows benchmarking the robustness and accuracy of SLAM and perception algorithms in the underground parking scenario.
    \item We evaluate several state-of-the-art semantic segmentation and SLAM algorithms on our dataset to show its practicability and challenging difficulties in the underground parking scenario.
    \item We open-source the SUPS dataset and the whole simulation underground parking scenario to enable researchers to make self-designed changes for specific tasks.
\end{itemize}

\section{RELATED WORK}
\label{II}
In this section, we first briefly review the mainstream benchmark datasets for autonomous driving tasks and then discuss the integration of the surrounding perception systems, semantic information and SLAM algorithms in the autonomous driving system.

\subsection{Autonomous driving datasets}
Existing autonomous driving datasets are diverse in carriers, environments, and sensors. However, most datasets were designed only for either SLAM or perception tasks. For example, EuRoC MAV\cite{Euroc}, TUM VI\cite{TUM}, and nuScenes\cite{nuScenes} did not provide label information for land-marking semantic segmentation tasks. Cityscapes\cite{Cityscapes}, nuScenes\cite{nuScenes}, ApolloScape\cite{apolloscape}, and BDD100k\cite{BDD100K} did not support depth estimation and SLAM tasks. Few researchers have worked on the combination of multiple tasks due to the extremely high cost of human labor and resources.

Only a few datasets contained underground parking scenarios. VIODE\cite{VIODE} provided underground scenes; however, the dataset was recorded with an unmanned aerial vehicle (UAV), which is not suitable for some autonomous driving algorithms. PS2.0\cite{PS2.0} is a dataset for parking slot detection tasks, which includes the underground parking scenario, but SLAM tasks are not supported. Thus, they have an obvious defect for underground automatic parking. At the same time, surround-view cameras are not applied in some mainstream autonomous driving datasets, such as KITTI\cite{kitti}, Cityscapes\cite{Cityscapes}, and BDD100k\cite{BDD100K}.

In summary, the open-sourced autonomous driving datasets were not sufficient for SLAM and perception tasks at the same time in the underground parking scenario. Taking advantage of the virtual scene, our dataset serves as a relatively complete solution to this issue by providing multiple sensors for diverse tasks, considering what a real underground parking scenario looks like. Tab.\ref{table_ex} provides the novelty of our dataset against existing datasets.

\subsection{Perception system for autonomous driving}
There is a growing interest in handling complex applications with 360° perception. GNNs~\cite{9046288,9744550} has been applied to improve parking slot detection at the sight of bird-eye-view (BEV)\cite{gcn}. Multi-pinhole cameras have been used in an end-to-end algorithm to solve motion planning tasks\cite{nvidia}. At the same time, perception algorithms such as semantic segmentation, detection, and scene encoding have caused heated discussion in SLAM tasks. Mask-SLAM joins a mask into the standard architecture to avoid disturbing information using semantic segmentation\cite{mask-slam}. DynaSLAM removed the dynamic objects with semantic segmentation\cite{dynaslam}. VIODE proposed the VINS-Mask which eliminated the use of feature points on some moving objects\cite{VIODE}. However, to the best of our knowledge, few open-source algorithms indeed make use of semantic information in feature extraction and frame matching. One of the obstacles to the development of such an algorithm is the lack of a dataset with plentiful well-labeled images for tracking and mapping. In addition, existing datasets mainly focus on labels of the drivable area, lane lines, and dynamic objects (such as other vehicles, cycles, and people); these features can hardly help track the location and build a usable map.
\section{OVERVIEW OF DATASET}
\label{III}
In this section, we introduce our dataset in detail with four sections. First, we describe our motivation to build this dataset. Then, we introduce the dataset generation procedure with the platform, simulator, carrier setups, and data processing method. Third, we explain the strategy to record the data sequence and split the set. Finally, we elaborate all contents of our dataset.

\subsection{Motivation}

\subsubsection{Surrounding perception}To finish the parking task, an automatic vehicle has to obtain the information of its surrounding environment without a blind corner. Our 360° perception system consists of four fisheye cameras surrounding the vehicle, a stereo pinhole camera set forward, and one LiDAR on the top. Fisheye cameras with a downside pitch angle are used for a larger FOV and better perception of ground signs.

\subsubsection{Multisensor for multitasks}SLAM tasks and various visual tasks are both challenging and indivisible from each other in the issue of autonomous parking. Semantic SLAM and multimodal sensing require many well-labeled frames as well as a complete set of sensors that cover vision, depth, ranging, and velocity measurements. The simulated multisensor system in the virtual scene provides GPS, IMU, LiDAR, depth measurements, and ground truth semantic segmentation aligned with every successive image according to timestamps. The detailed configuration of sensors is introduced in subsection \ref{sub_b}.

\subsubsection{Changeable simulated scene}Our simulated underground parking scenario was built based on the Unity Engine\cite{unity}. Users can easily modify the scene and rebuild it for the specific application. For example, the parking lot structure, traffic signs, illumination, semantic labels, and travel routes can all be changed. An overview of the existing scene is described in subsection \ref{sub_b}.

\subsection{Dataset generation}
\label{sub_b}

The SUPS dataset generation procedure involves three steps: virtual scene construction, simulator setups, recording and processing. These steps were completed on a computer with Ubuntu 18.04, Intel(R) Core i7-11800H CPU, and a GeForce RTX 3060 (Laptop) GPU.

\begin{figure}[t]
\centering
\includegraphics[width=8cm]{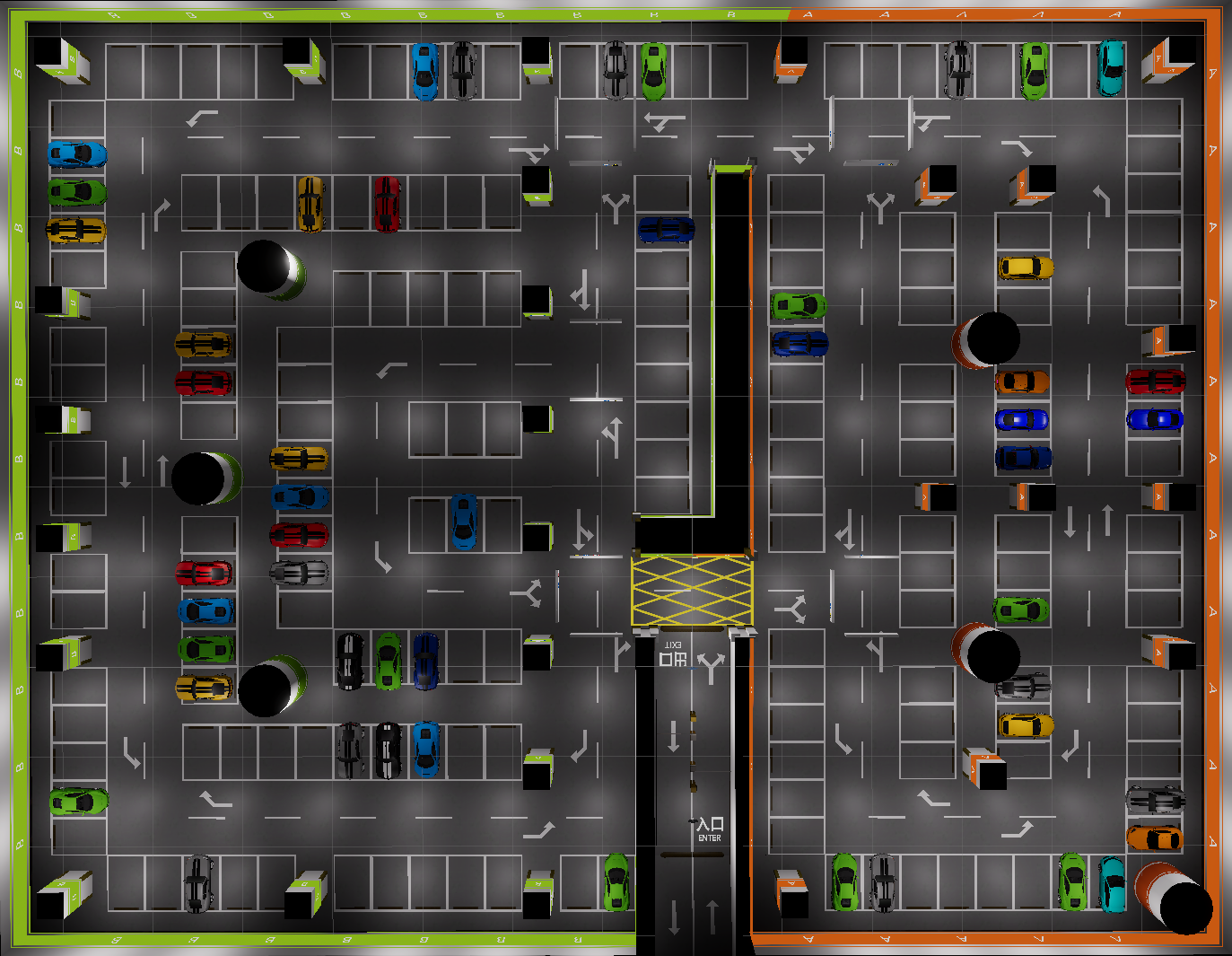}
\caption{An overview of the simulating scene. \textbf{Area A} is the orange part shown on the right side and \textbf{Area B} is the green part shown on the left side.}
\label{scene_overview}
\end{figure}

\subsubsection{Virtual scene construction}We built our 3D simulating scene for the underground parking scenario through the Unity Engine\cite{unity}. The whole scene consists of two areas \textbf{Area A} and \textbf{Area B} as shown in Fig. \ref{scene_overview}. \textbf{Area A} occupies $2635 m^{2}$ with a road length of $120 m$. \textbf{Area B} occupies $2015 m^{2}$ with $195 m$ long road. The drivable road in the whole scene is approximately $350 m$ long, and the parking slot occupies an area of $4650 m^{2}$. There are several different road loops in the scene for providing loop closure constraints in the SLAM task. To enrich the scene, we involve items such as walls, pillars, static vehicles, parking slots, lane lines, drivable areas, collision avoidance strips, speed bumps, arrows, and road signs. Additionally, there are challenging scenarios for perception and SLAM tasks in the simulated scene, similar to the real-world underground parking scenarios, such as crossing, entrance, one-way street, dazzle, dim area, and low-texture area(see Fig. \ref{full_sensor}). Furthermore, to maximally restore the actual underground parking scenario, we use spotlights at the ceiling.

\subsubsection{Simulator setups}
\begin{figure}
\centering
\includegraphics[width=8cm]{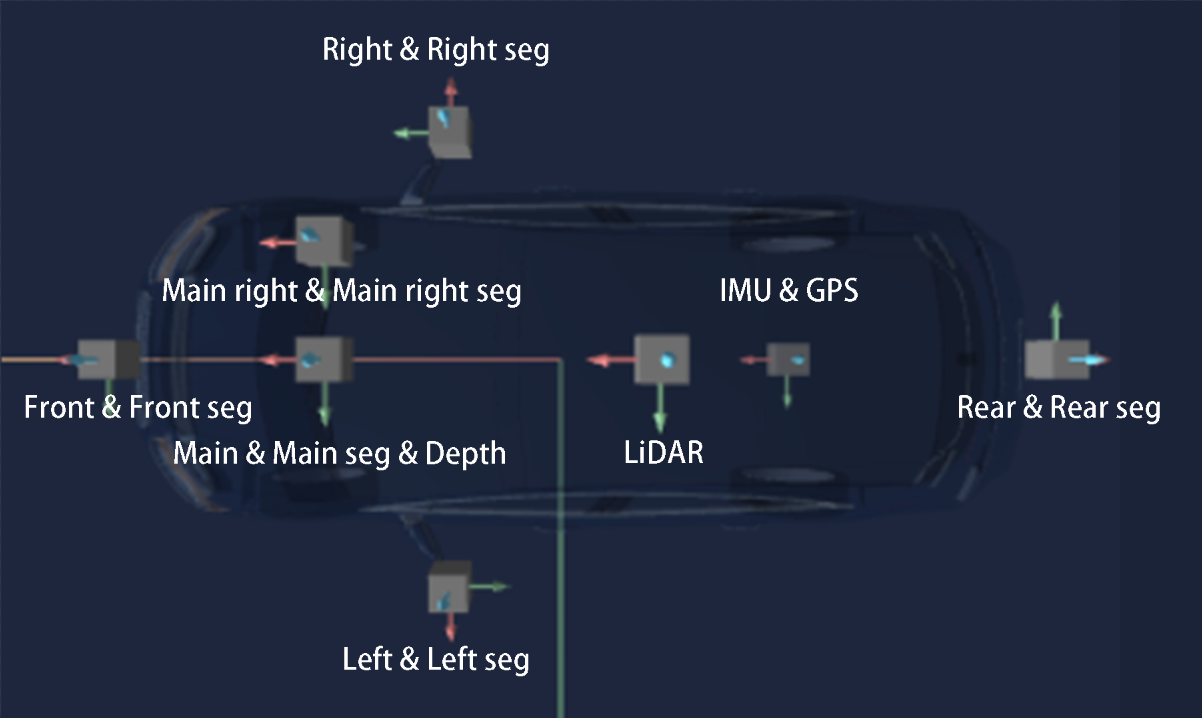}
\caption{Positions of sensors on ego vehicle. The names of the cameras are written in short. For example, \textit{Front \& Front seg} means \textit{Front camera and Front semantic segmentation camera}.}
\label{sensor_on_car}
\end{figure}

On the LGSVL\cite{lgsvl} platform, we create a Lincoln2017NKZ autonomous vehicle model and apply multiple sensors (detail of sensors shown in Tab. \ref{sensor_detail_table}). An overview of sensors on the ego car is shown in Fig. \ref{sensor_on_car}. We also provide all intrinsic and extrinsic parameters for the sensors. To obtain the ground truth of the semantic segmentation task, we install six additional semantic segmentation cameras on the ego model that share the same parameters as the corresponding four fisheye cameras and two pinhole cameras. Fig. \ref{full_sensor} shows an example of observations and measurements from all sensors.

\begin{table}[t]
\caption{Detail of sensors in the ego vehicle model}
\label{sensor_detail_table}
\begin{center}
\begin{tabular}{|c||c||c|}
\hline
Sensor name & Position & Sampling Frequency\\
\hline
Fisheye camera 1 & front &  20 Hz\\
\hline
Fisheye camera 2 & left &  20 Hz\\
\hline
Fisheye camera 3 & right &  20 Hz\\
\hline
Fisheye camera 4 & rear &  20 Hz\\
\hline
Pinhole camera 1 & top & 20 Hz\\
\hline
Pinhole camera 2 & top-right & 20 Hz\\
\hline
Depth camera & top & 20 Hz\\
\hline
GPS Odometry sensor & rear axle center & 30 Hz \\
\hline
IMU & rear axle center & 100 Hz \\
\hline
LiDAR & top-behind & 7 Hz(rotation)\\
\hline
\end{tabular}
\end{center}
\end{table}

\subsubsection{Data processing}
\begin{figure}
    \centering
    \includegraphics[height=2.6cm]{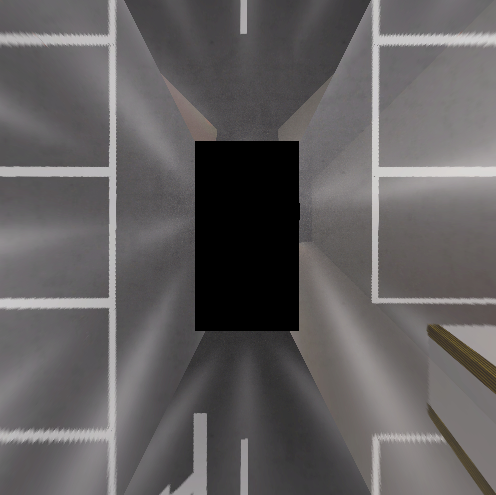}
    \includegraphics[height=2.6cm]{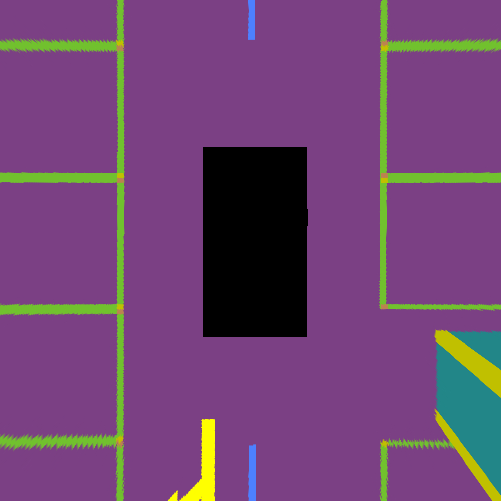}
    \includegraphics[height=2.6cm]{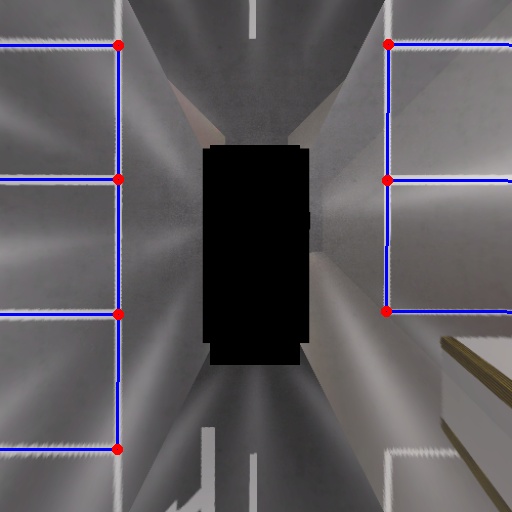}
    \caption{Example of the parking slot detection data form. The left is an original image, the medial is the ground truth, and the right shows the GCN\cite{gcn} form description.}
    \label{gcn_example}
\end{figure}

\begin{figure}[t]
    \centering
    \includegraphics[width=8cm]{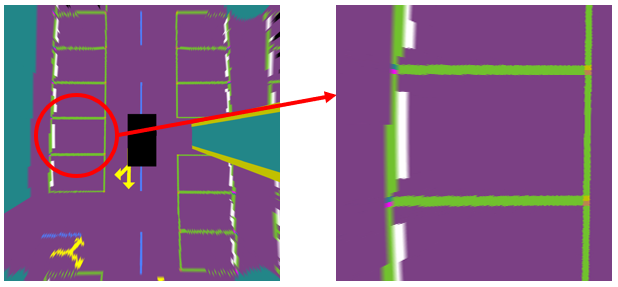}
    \caption{The description of a single parking slot. Four corners of the parking space are labeled with different colored squares, which show the width of parking lines and the order of corners. The center line, object-level mask, and corners with vectors can be generated with these labels easily.}
    \label{parking_slot}
\end{figure}

Only a few existing datasets provide enough information for parking slot detection. In the real world, it is common that parking lines are worn down or hid by cars and obstacles, which makes it difficult for human laborers to label them accurately. In most datasets, the FOV of the BEV cannot cover a complete parking space along the roadside. At the same time, different algorithms described the parking space variously. For instance, GCN-parking-slot described a parking space with two corners at the entry side and the correspondence of corners\cite{gcn}. Our dataset provides an adaptable description of the parking slots, which includes all the information to separate a parking space (see Fig. \ref{parking_slot}). Corners with vectors, center lines, and object-level masks can be easily extracted with ordered point coordinates. For instance, we provide data for parking slot detection using the description form from GCN\cite{gcn}(see Fig. \ref{gcn_example} for details).

Additionally, since bird-eye-view(BEV) images are especially helpful in perception and decision making when operating automatic parking, besides the original camera frames, we provide BEV images and segmentation ground truth (see Fig. \ref{origin_seg}) projected by the inverse perspective mapping(IPM) method.

\subsection{Acquisition strategy}

\subsubsection{SLAM tasks}
\begin{figure}[t]
\centering
\subfigure[Loop A and Loop B]{
\includegraphics[width=8cm]{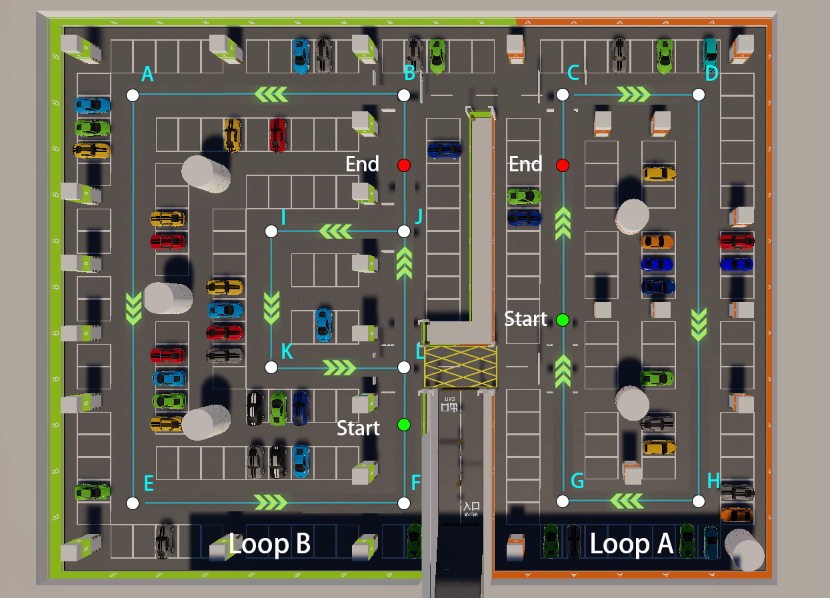}
}
\subfigure[Loop C]{
\includegraphics[width=8cm]{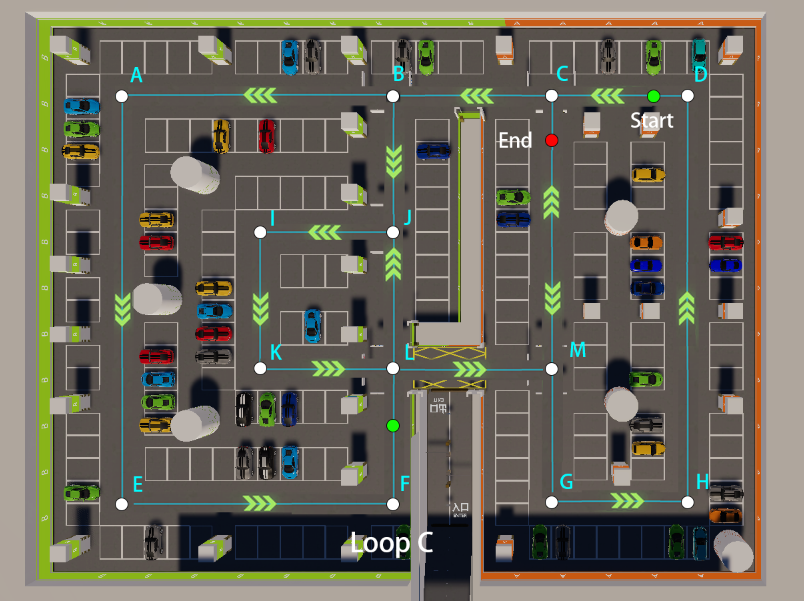}
}
\caption{Overview of three routes and driving sequences. The green point is where the ego starts, and the red point is where the ego stops. Each white point at the crossing is labeled with a capital letter from $A$ to $M$ to describe the route when ego car passes through. As (a) shows, in \textbf{Loop A}, ego car drives through $Start \to C \to D \to H \to G \to End$. In \textbf{Loop B}, ego car drives through $Start \to B \to A \to E \to F \to J \to I \to K \to L \to End$. As (b) shows, in \textbf{Loop C}, ego car drives through $Start \to C \to G \to H \to D \to B \to L \to M \to C \to B \to F \to E \to A \to B \to J \to I \to K \to M \to End$.}
\label{route}
\end{figure}

Three routes at two speed levels are recorded in our dataset. Fig. \ref{route} shows three driving routes (\textbf{Loop A}, \textbf{Loop B} and \textbf{Loop C}) sequence in detail. \textbf{Loop A} has a single loop closure with a route approximately $160 m$ long, \textbf{Loop B} has two loops ($60 m$ long and $150 m $ long) with a route approximately $250 m$. \textbf{Loop C} contains several complex loops and the route is approximately $700 m$ as shown in Fig. \ref{route}. The speed limit on most underground parking scenarios is $5 km/h$, so the high-speed level of our recorded data is $5 km/h$, and the car drives at $3.5 km/h$ at a low-speed level. Additionally, considering that some SLAM algorithms (such as Vins-MONO\cite{vins-mono}) need enough IMU excitation when initializing, we provide an extra version for each data record that has an initialization process, in which we make some slow turns before the ego car drives according to the route.

\subsubsection{Perception tasks}We generate the depth images, original frames, and their corresponding semantic segmentation ground truth from a record of driving in Loop C, which covers almost all information in the underground parking scenario. Captured frames are split into three subsets at a ratio of 7:1:2 as the training, validation, and testing sets. The training set serves for training perception models only, the validation set can be used as a supplement of the training set or for model selection, and the testing set is for model evaluation only.

\subsection{Dataset content}

We record our dataset in rosbag format. Stereo cameras and surrounding cameras are partially activated with other sensors because long periods and quantities of sensors are costly for memory and computation. However, we still provide bags in which all sensors described in subsection \ref{sub_b} are activated when the vehicle travels in \textbf{Loop A} and \textbf{B}. Records are named as $Route\_SpeedLimit\_InitializeProcess\_Sensor.bag$. 

\begin{itemize}
    \item \textit{Route} describes in which loop the bag file is recorded. Candidates are \textit{Loop A}, \textit{Loop B}, \textit{Loop C}.
    \item \textit{SpeedLimit} describes the average traveling speed. Candidates are \textit{fast}, \textit{slow}.
    \item \textit{InitializeProcess} describes whether there is an extra process that helps initialize IMU. Candidates are \textit{disturbed} (for sure), \textit{direct} (for not).
    \item \textit{Sensor} describes the sensors activated during records. \textit{Stereo} means stereo forward cameras and other types of sensors are activated. \textit{Surround} means surrounding fisheye cameras and other types of sensors are activated. \textit{Full} means all sensors are activated.
\end{itemize}

\begin{figure}
\centering
\subfigure[Front view]{
    \includegraphics[width=4cm]{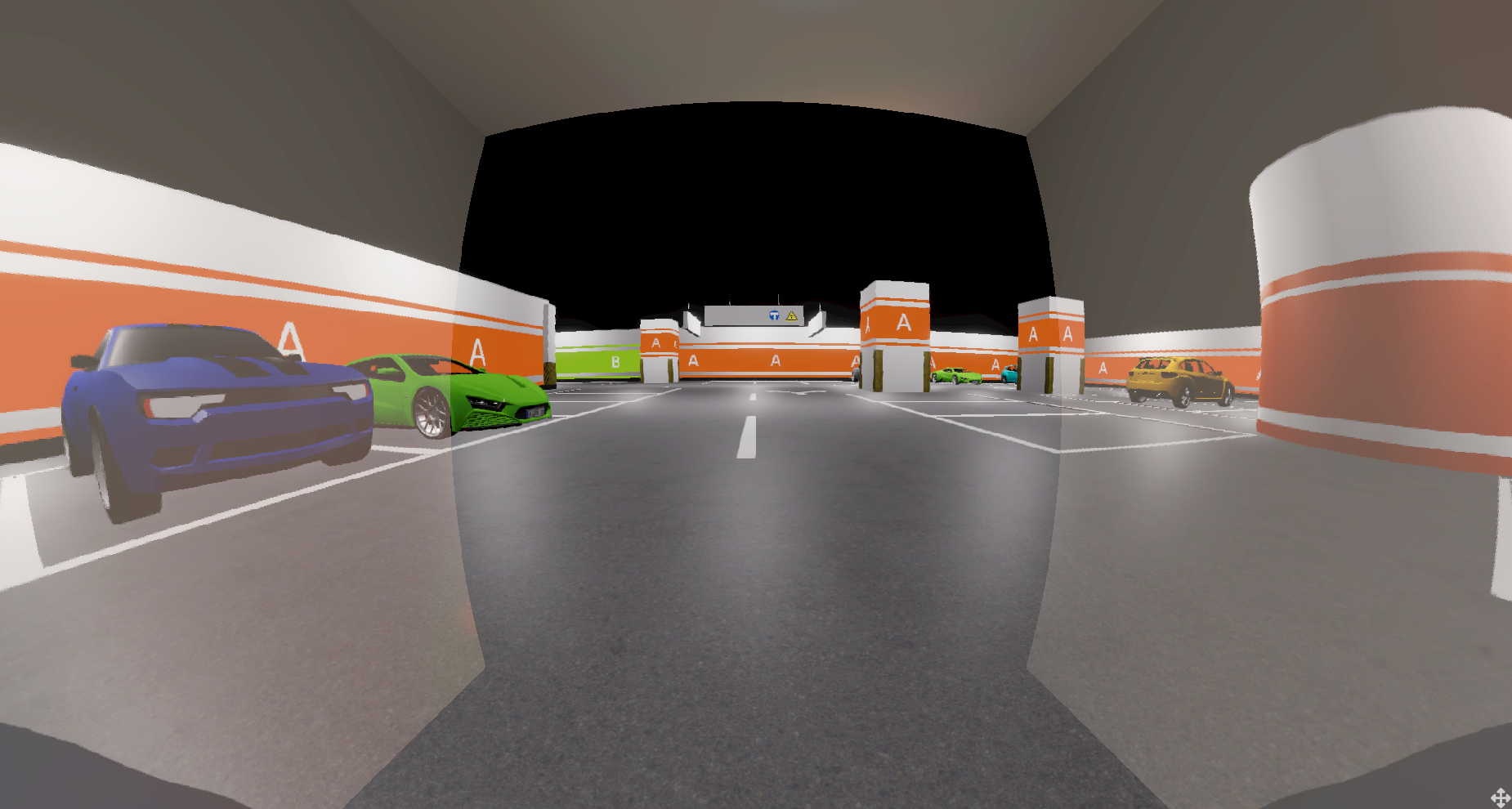}\vspace{6pt}
    \includegraphics[width=4cm]{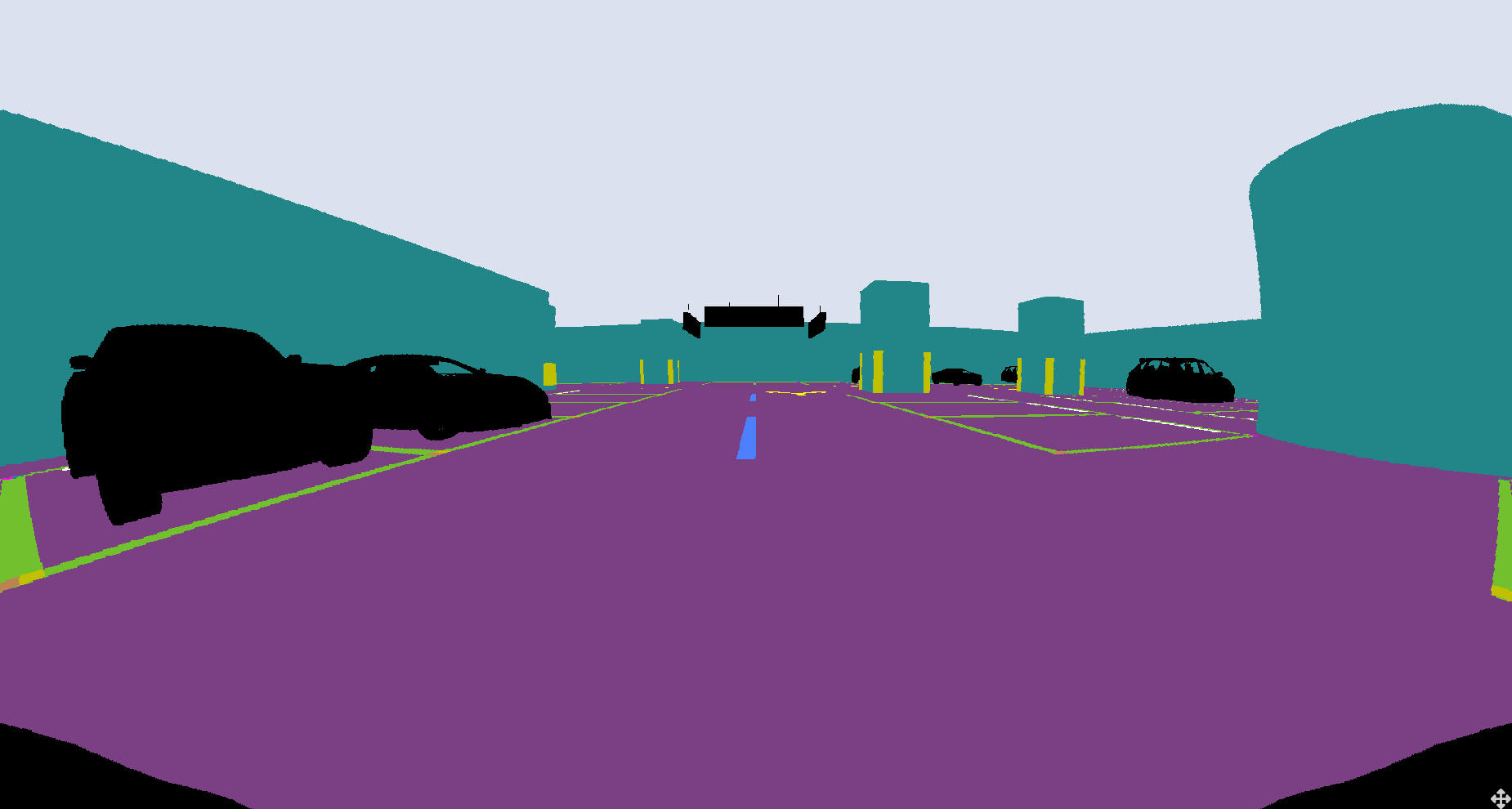}\vspace{6pt}
}
\subfigure[Bird-eye view]{
    \includegraphics[width=4cm]{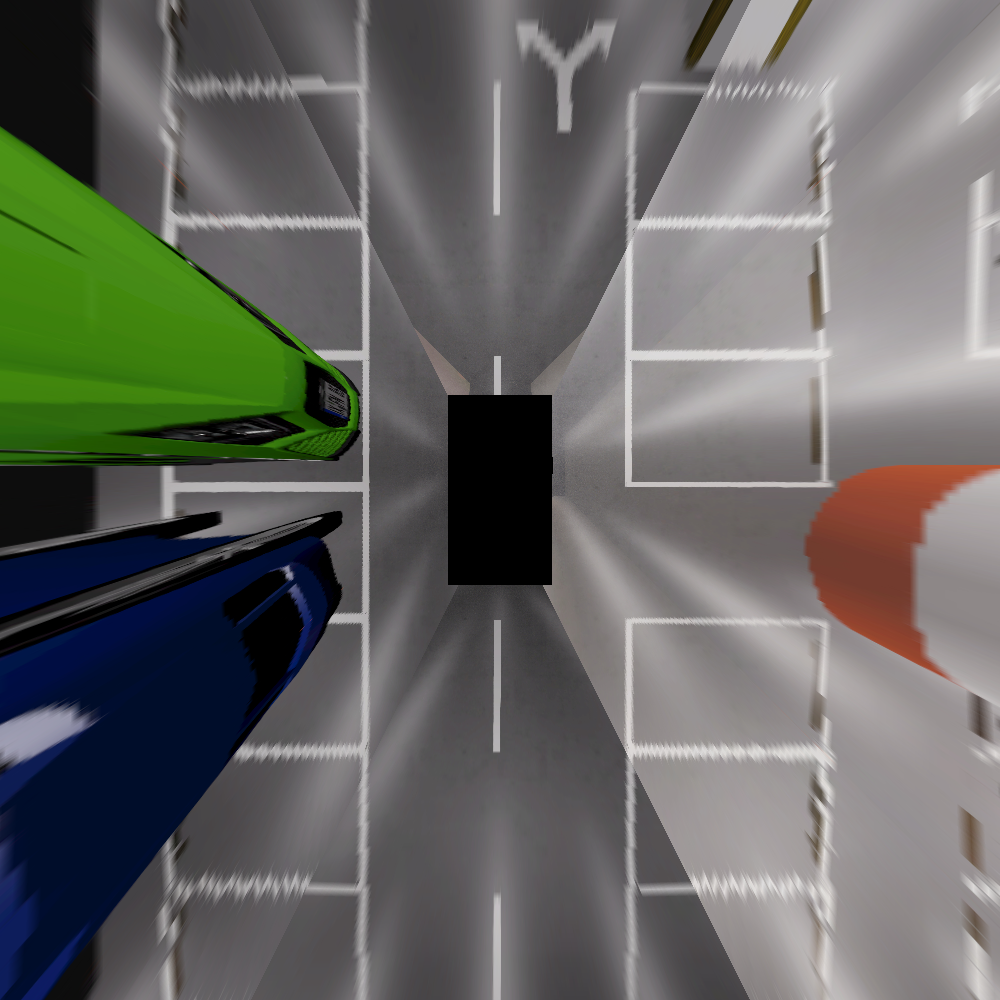}\vspace{6pt}
    \includegraphics[width=4cm]{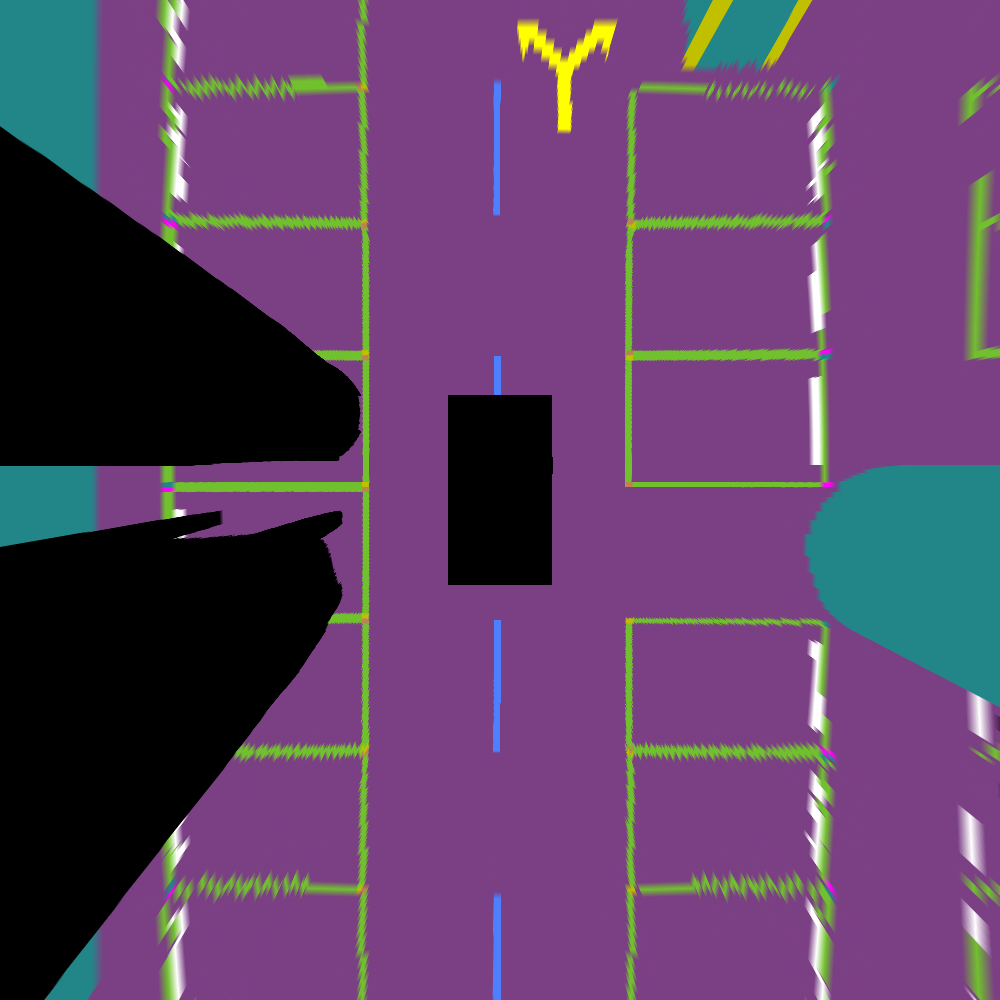}\vspace{6pt}
}
\caption{Comparison of images and semantic segmentation ground truth in front view and BEV. Items in the scene are labeled separately: drivable area, wall, pillar, static vehicle, parking lines, lane lines, collision avoidance strips, speed bumps, arrows. The left column shows the original image in the front view and the projected BEV, and the right column shows the ground truth.}
\label{origin_seg}
\end{figure}

More than 5,000 frames are provided with ground truth for supported perception tasks, such as semantic segmentation, parking slot detection, and depth estimation. Details for some supported tasks will be introduced in section \ref{IV}. Fig. \ref{origin_seg} shows an example of classification labels in our dataset.

\section{Experiments}
\label{IV}
\subsection{Semantic segmentation}

\begin{figure}
    \centering
    \includegraphics[height=2cm]{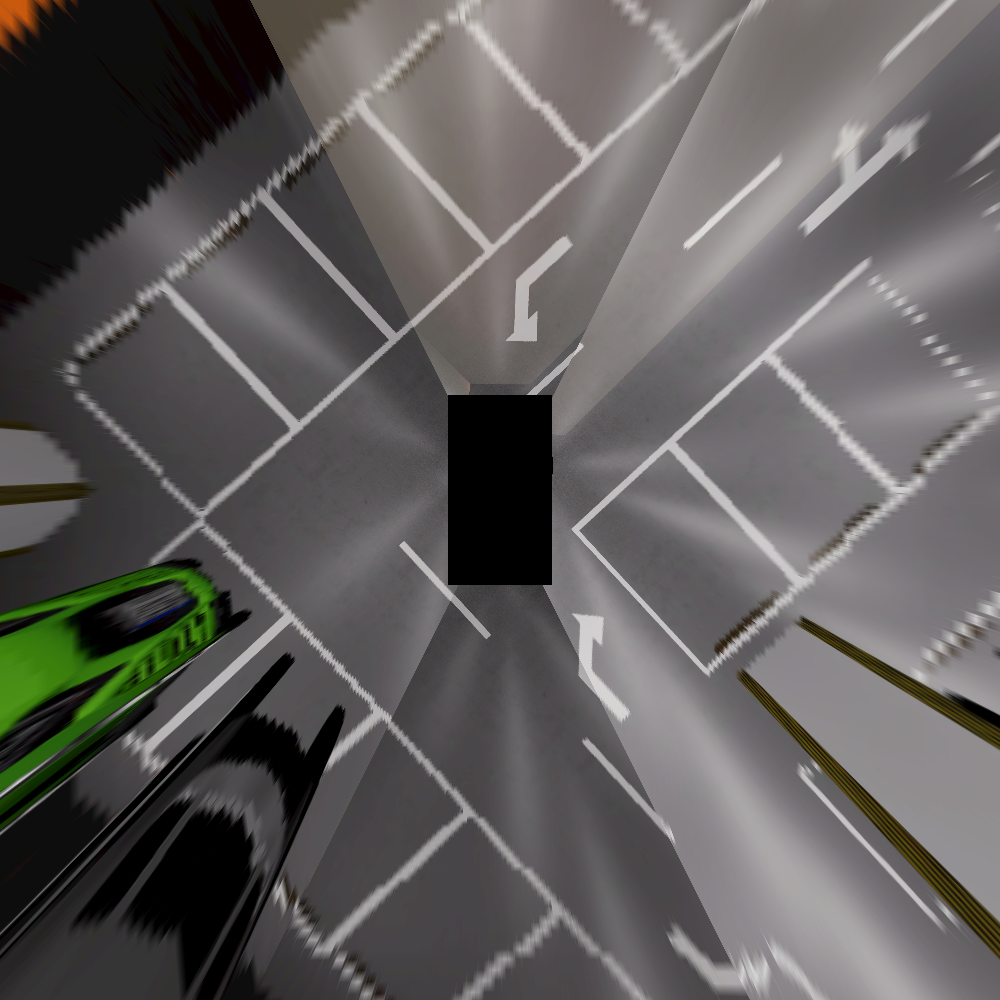}
    \includegraphics[height=2cm]{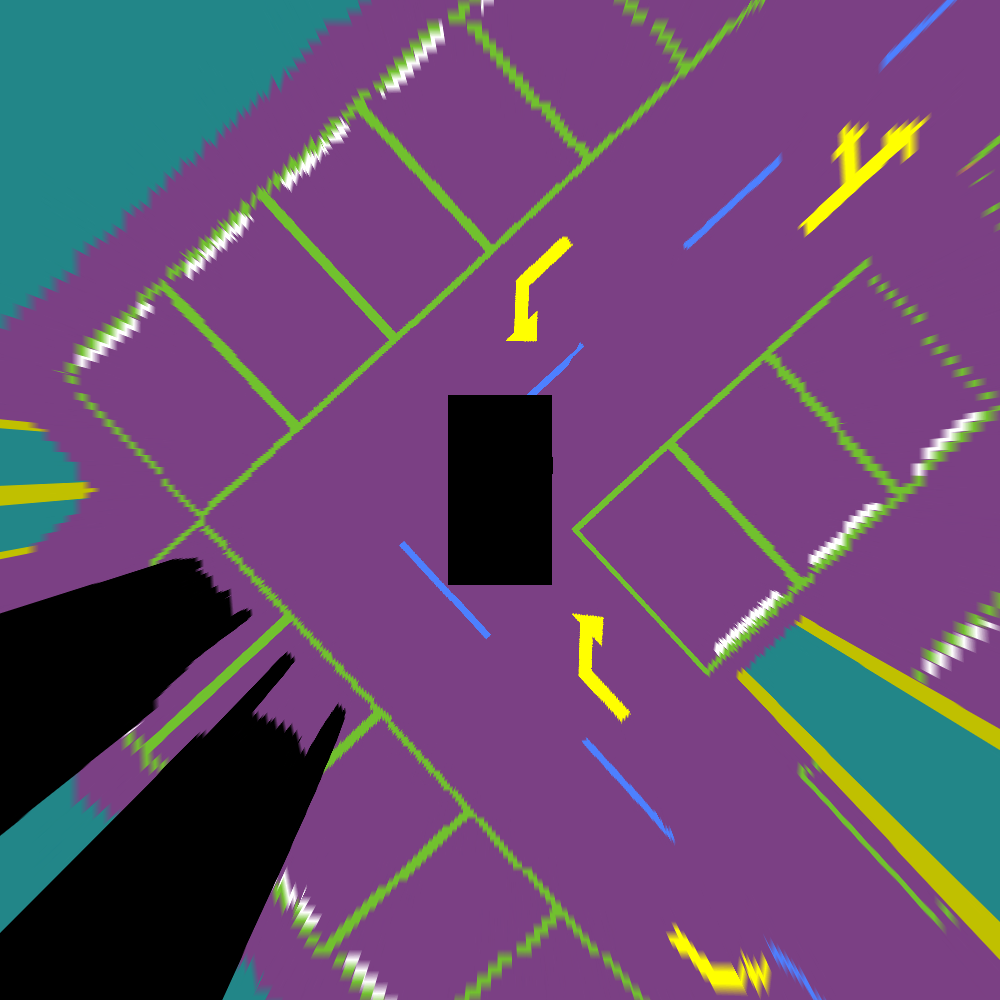}
    \includegraphics[height=2cm]{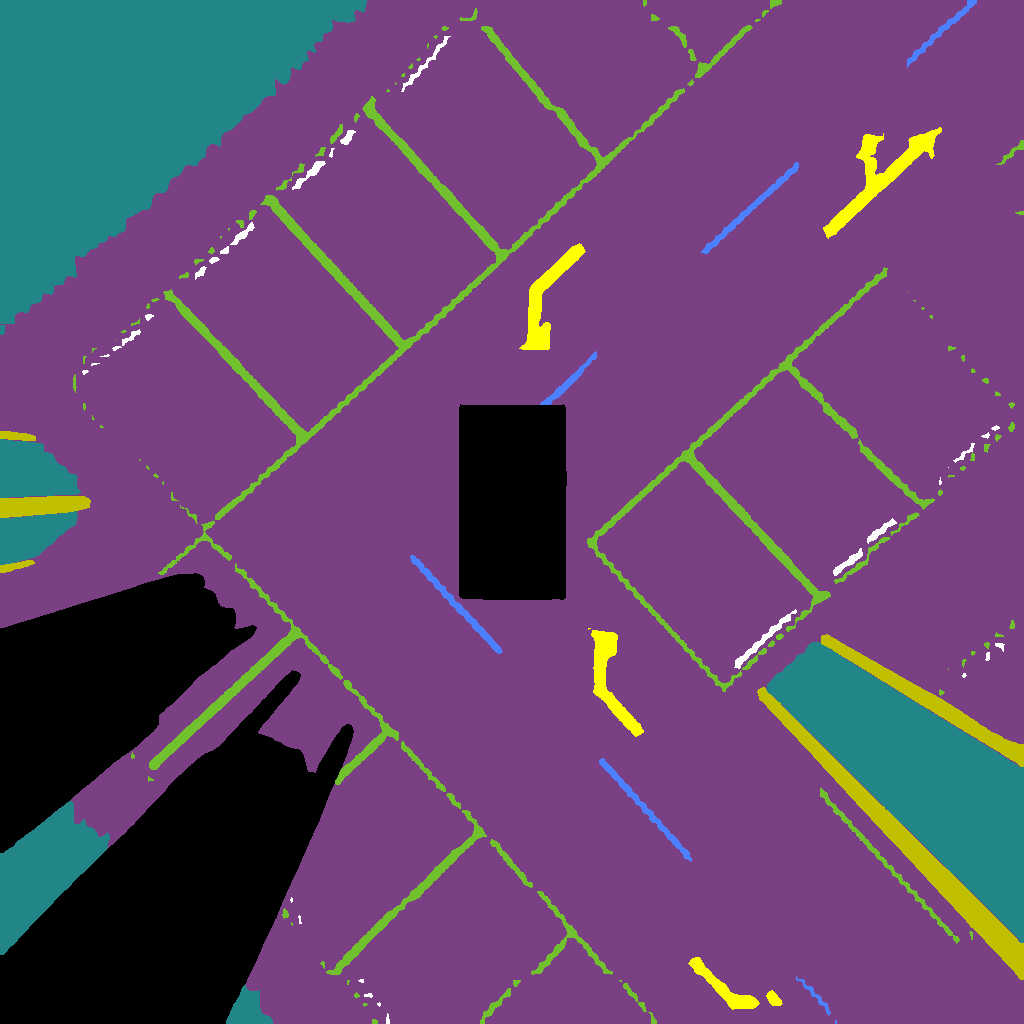}
    \includegraphics[height=2cm]{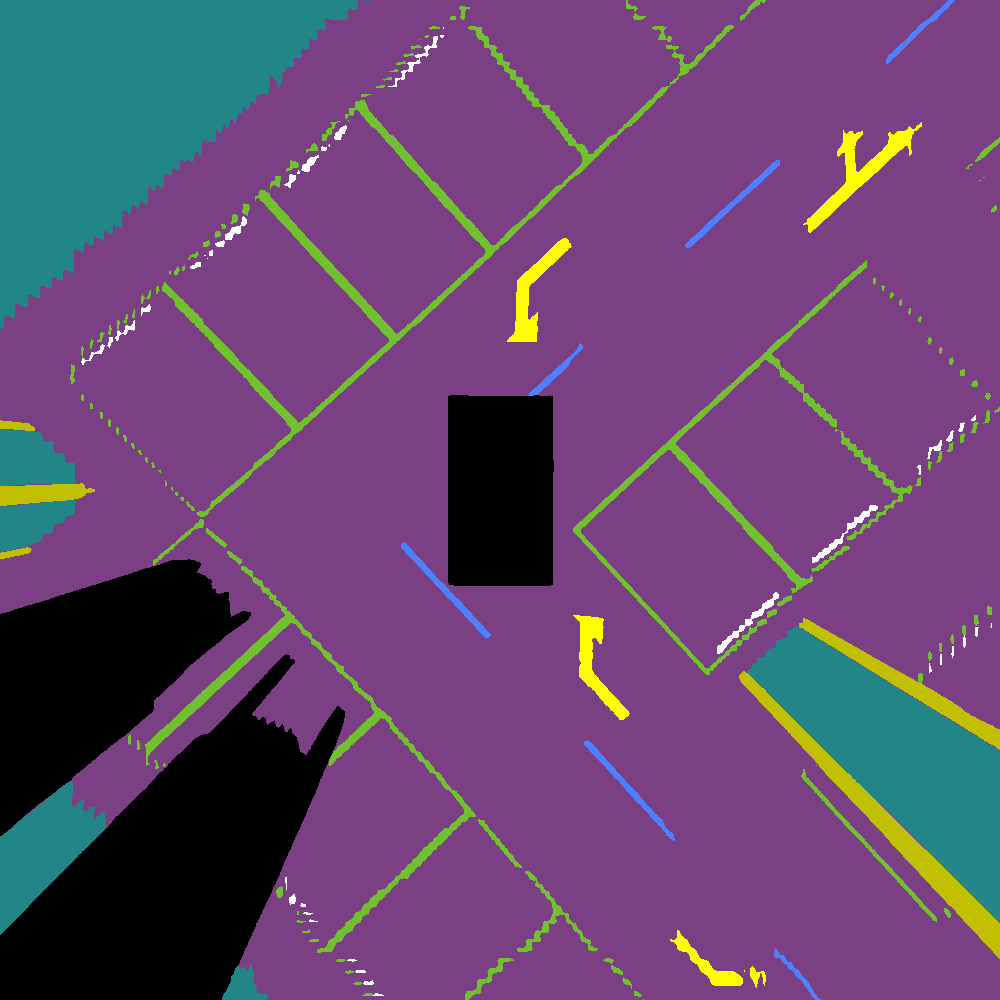}\\
    \vspace{2pt}
    \includegraphics[height=2cm]{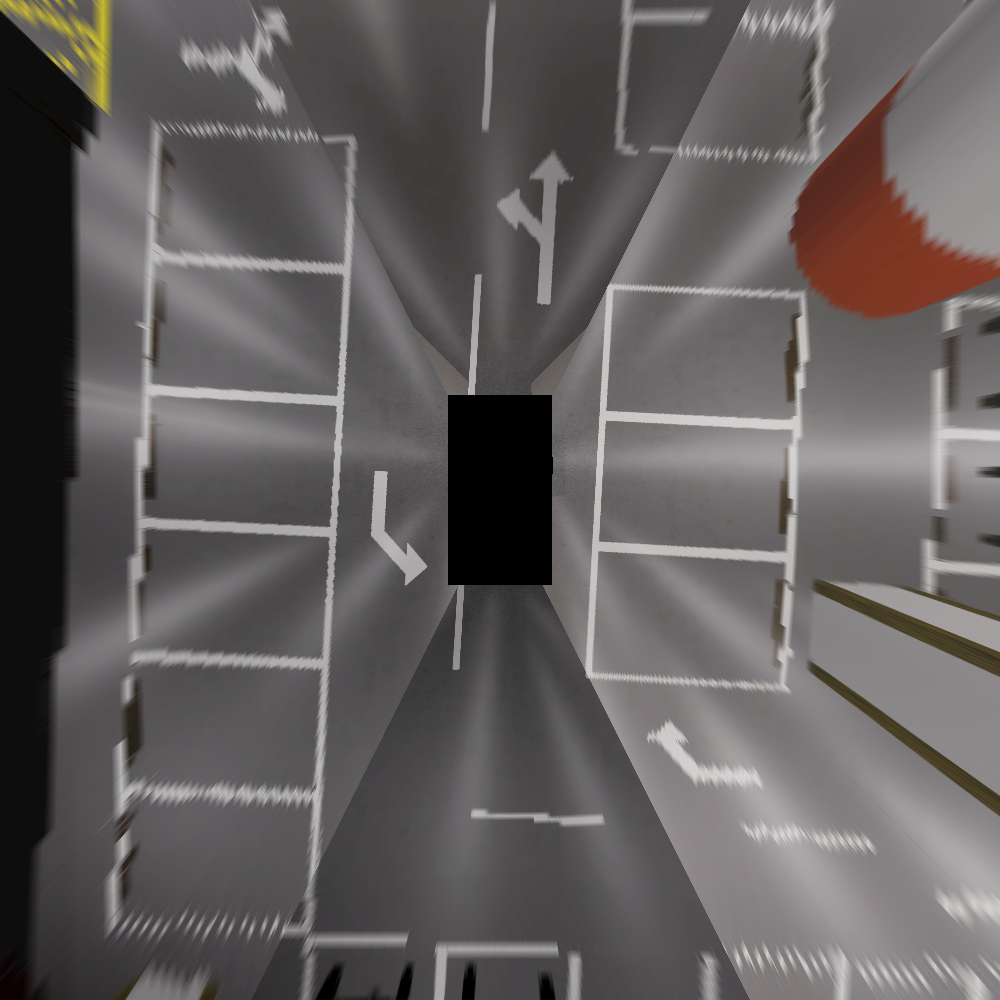}
    \includegraphics[height=2cm]{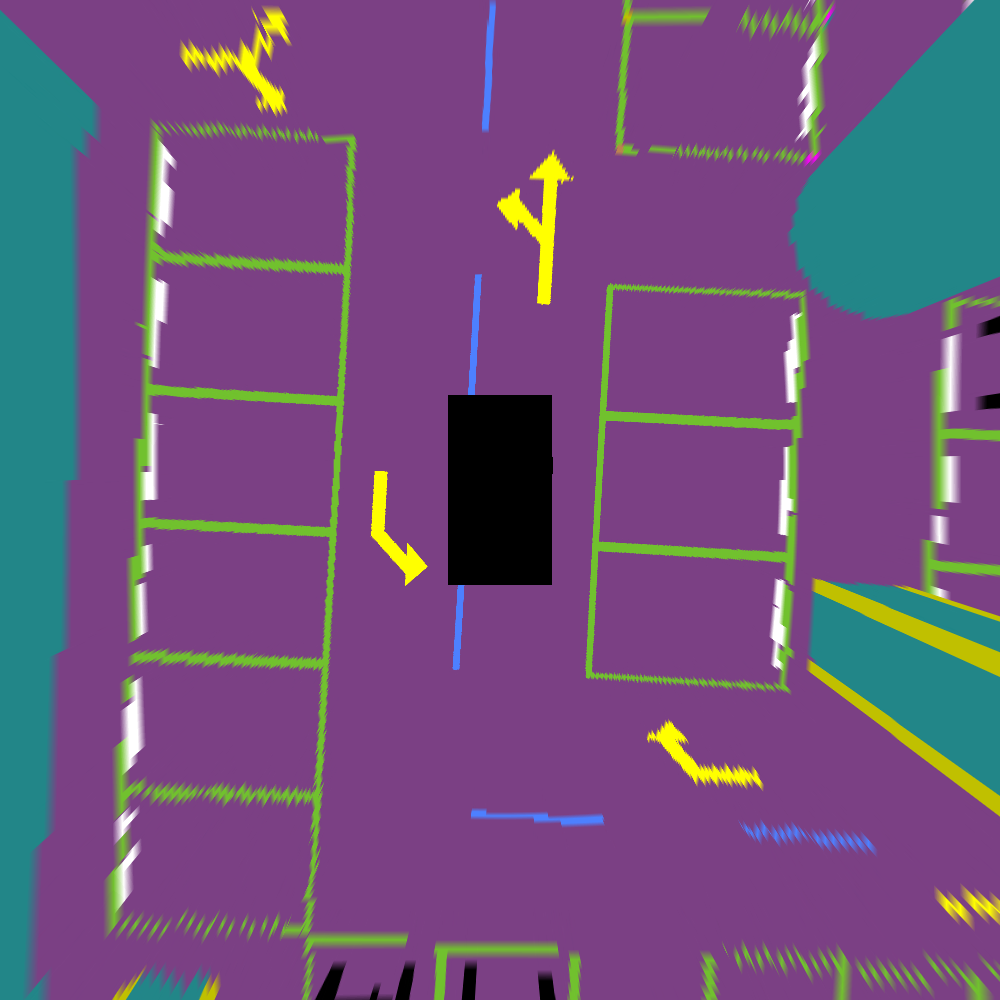}
    \includegraphics[height=2cm]{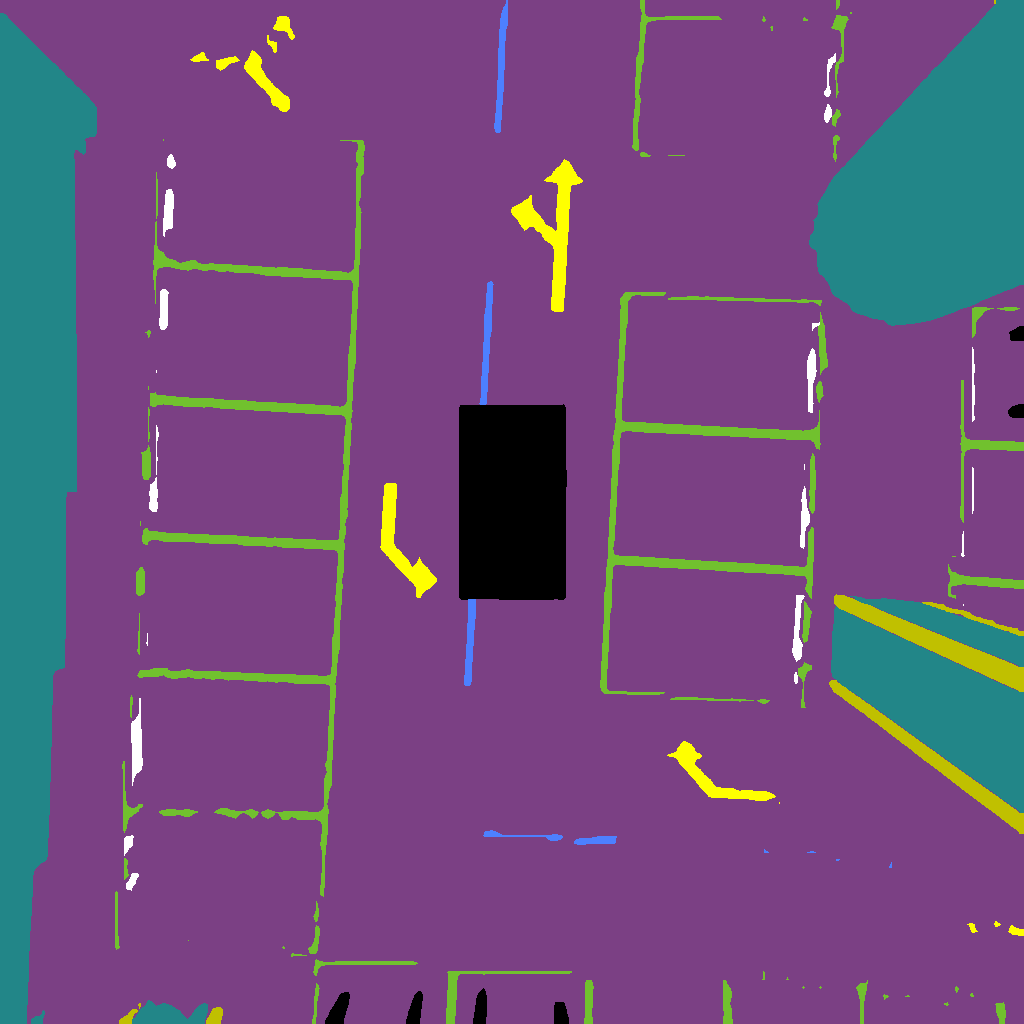}
    \includegraphics[height=2cm]{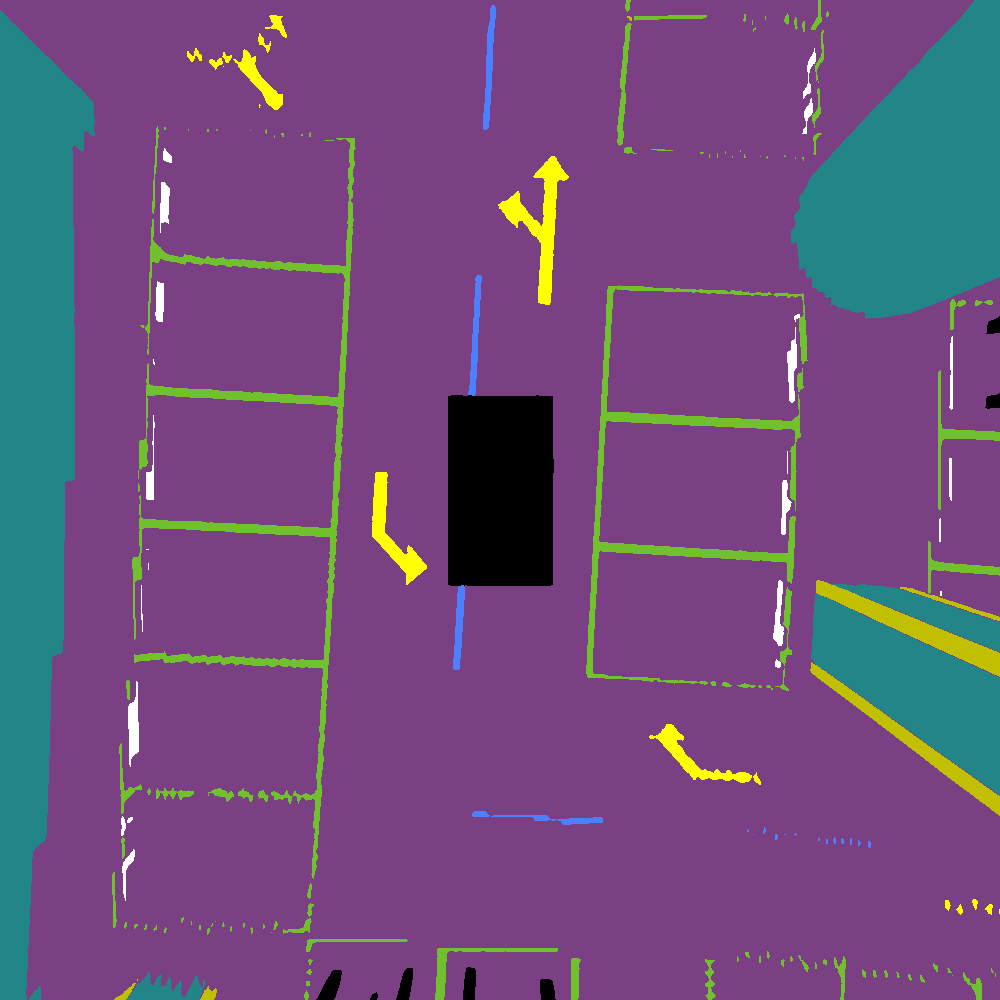}\\
    \vspace{2pt}
    \includegraphics[height=2cm]{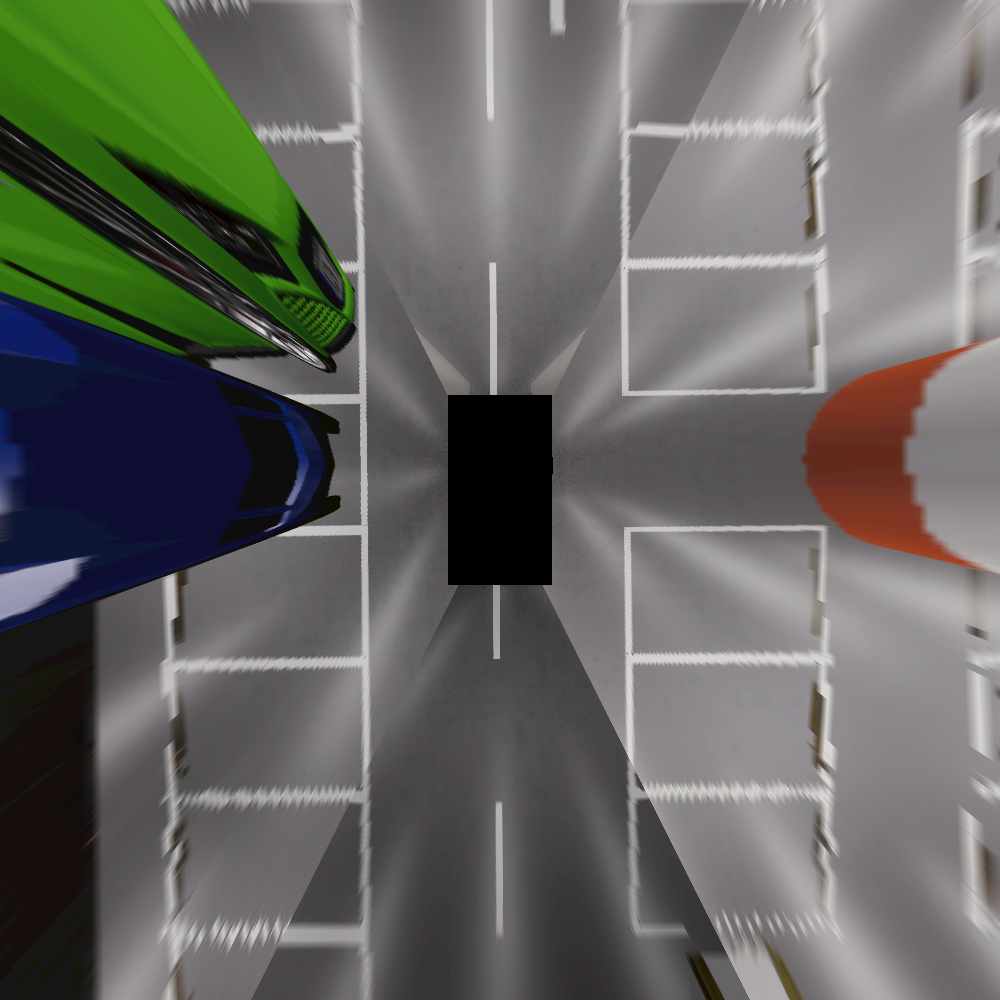}
    \includegraphics[height=2cm]{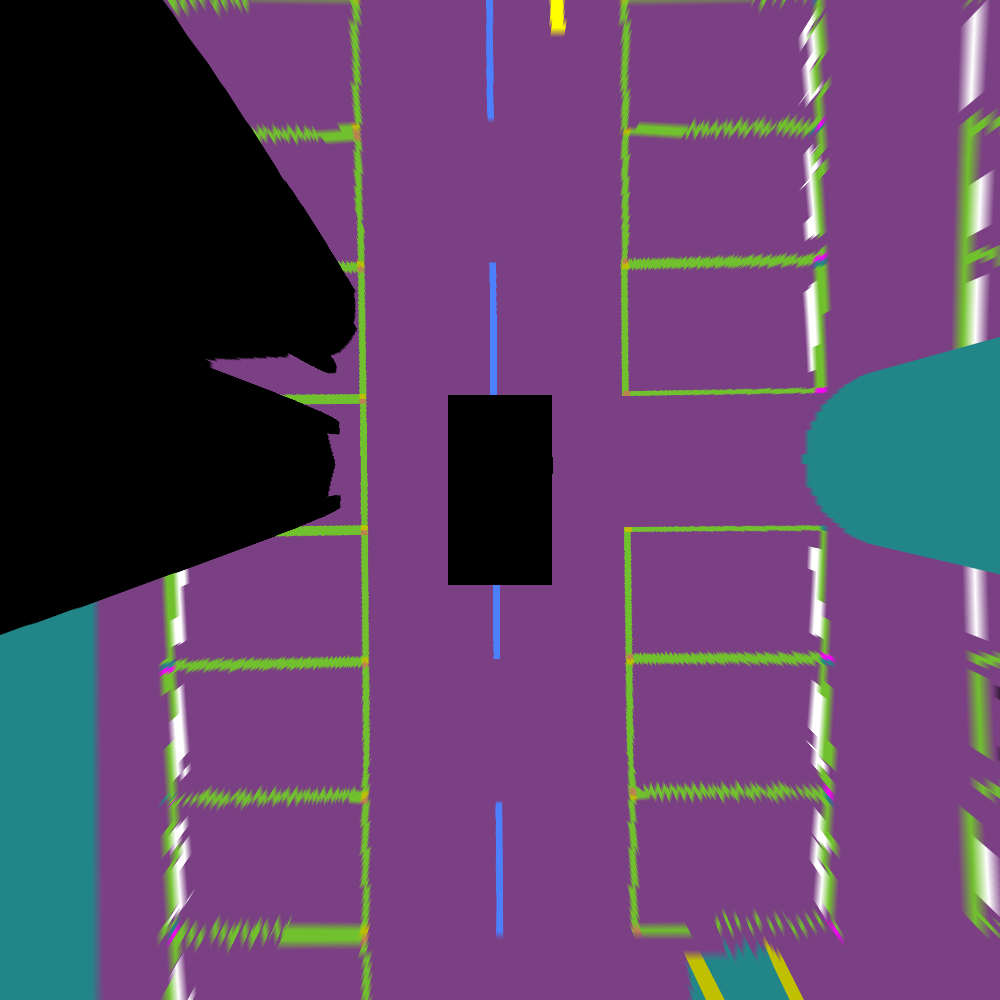}
    \includegraphics[height=2cm]{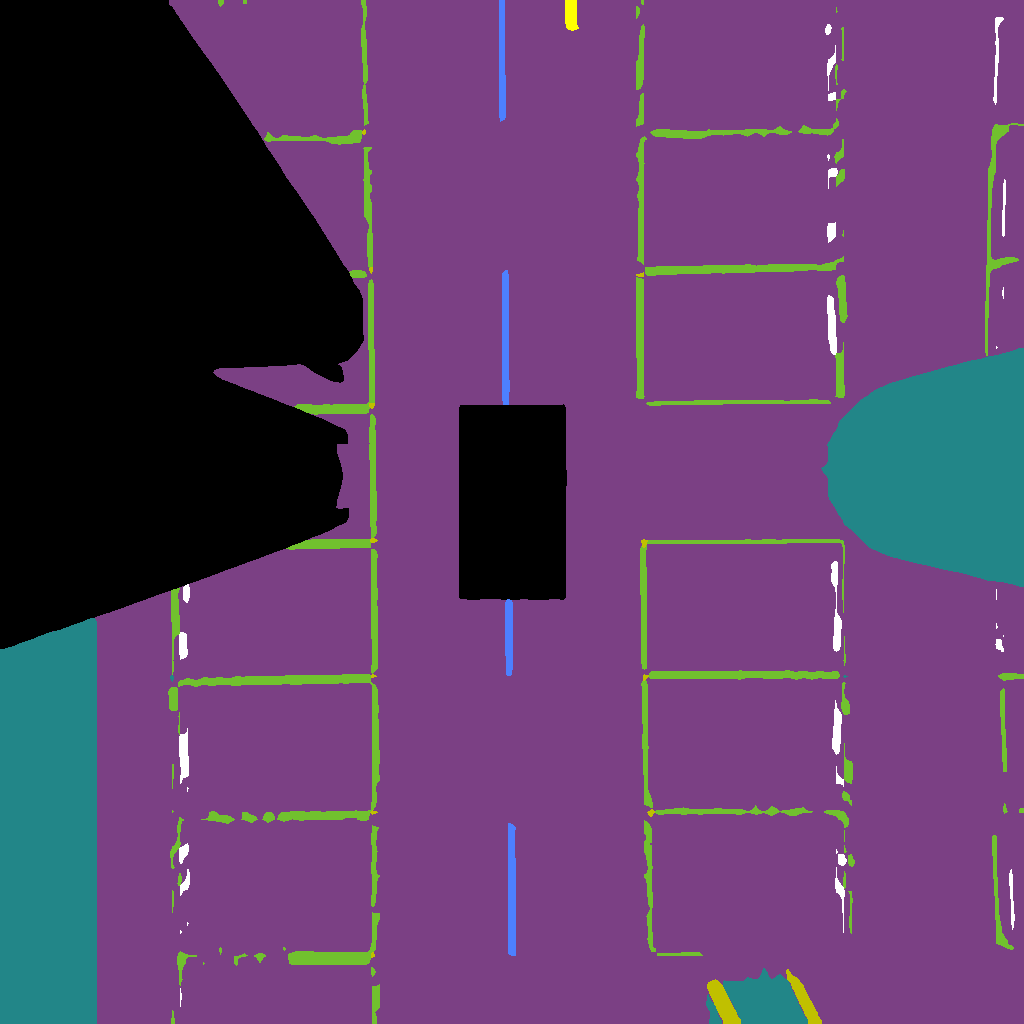}
    \includegraphics[height=2cm]{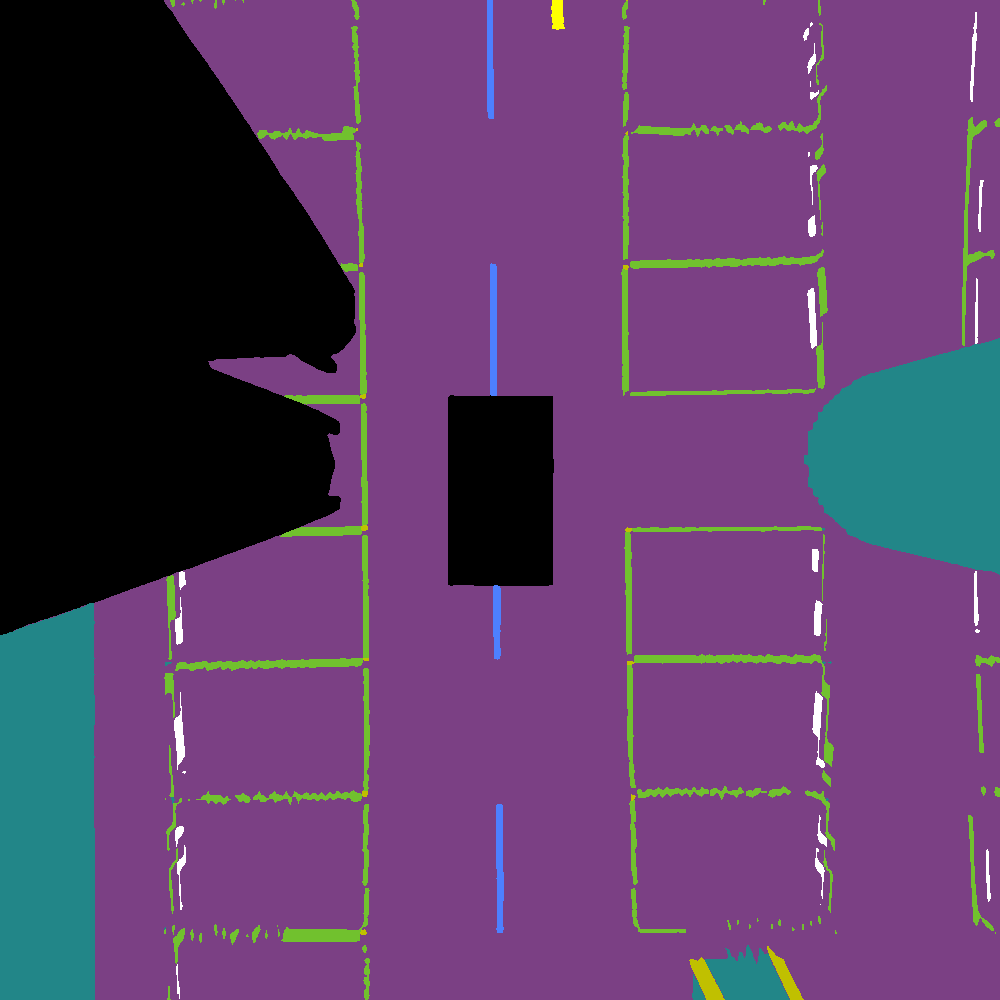}\\
    \vspace{2pt}
    \includegraphics[height=2cm]{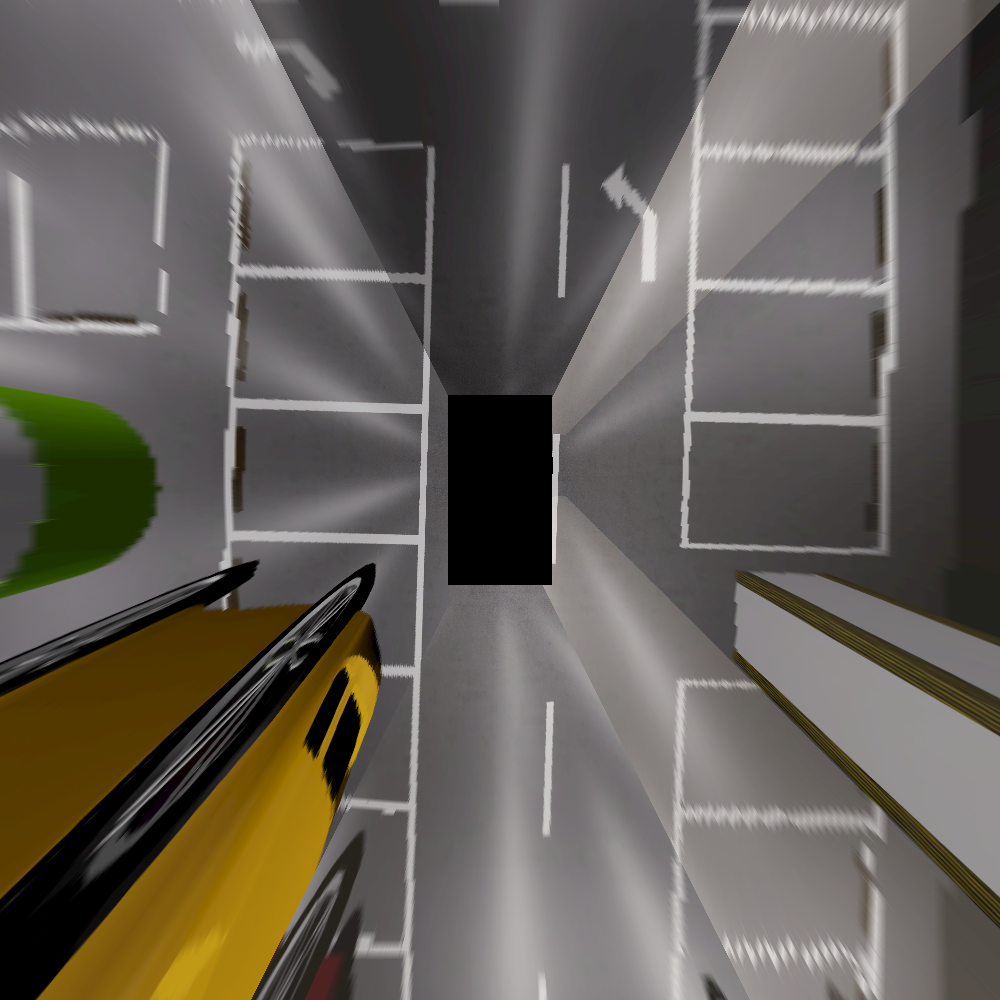}
    \includegraphics[height=2cm]{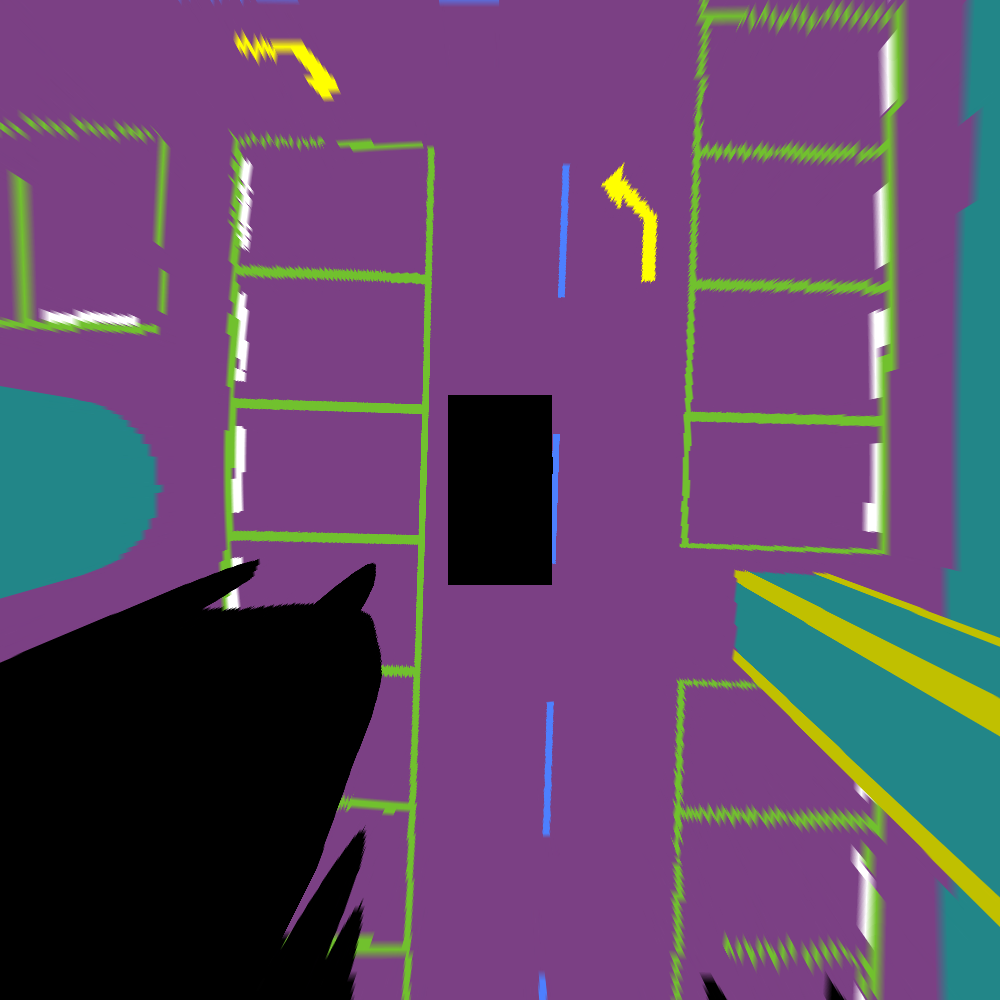}
    \includegraphics[height=2cm]{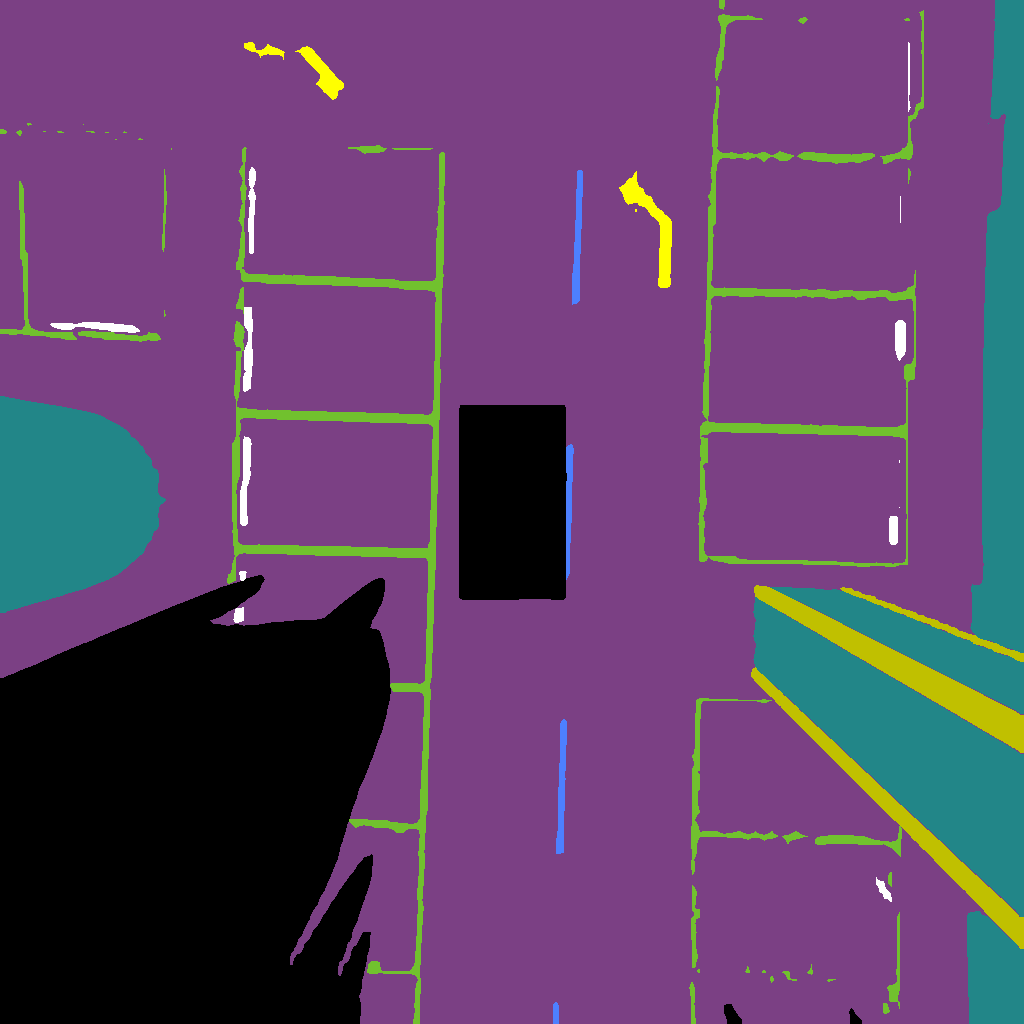}
    \includegraphics[height=2cm]{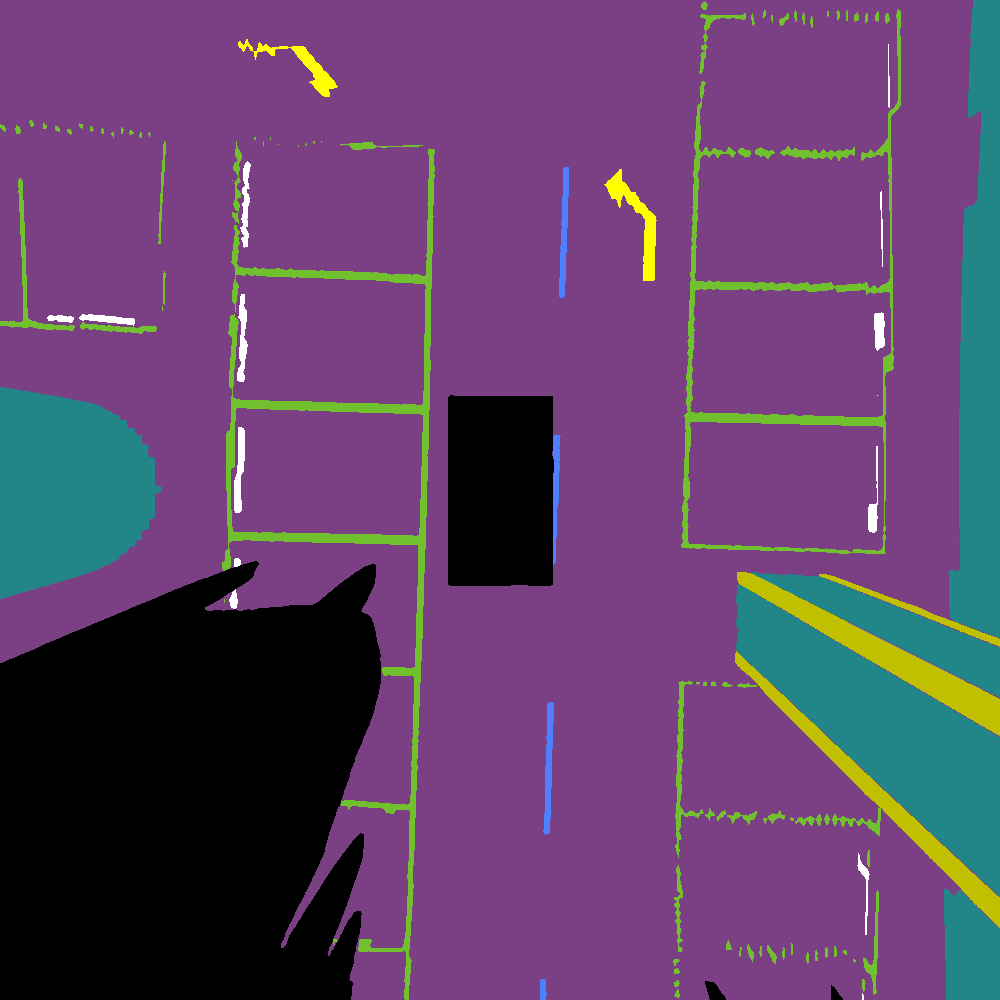}\\
    \vspace{2pt}
    \includegraphics[height=2cm]{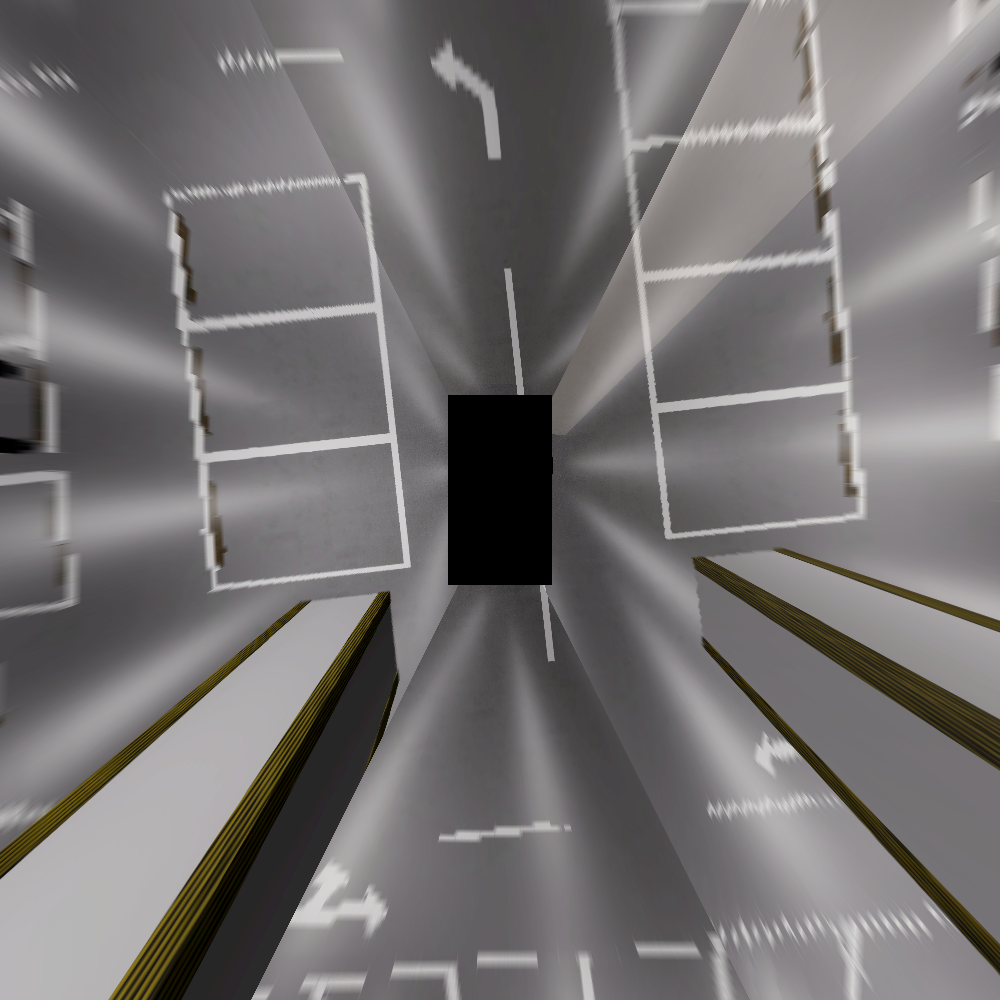}
    \includegraphics[height=2cm]{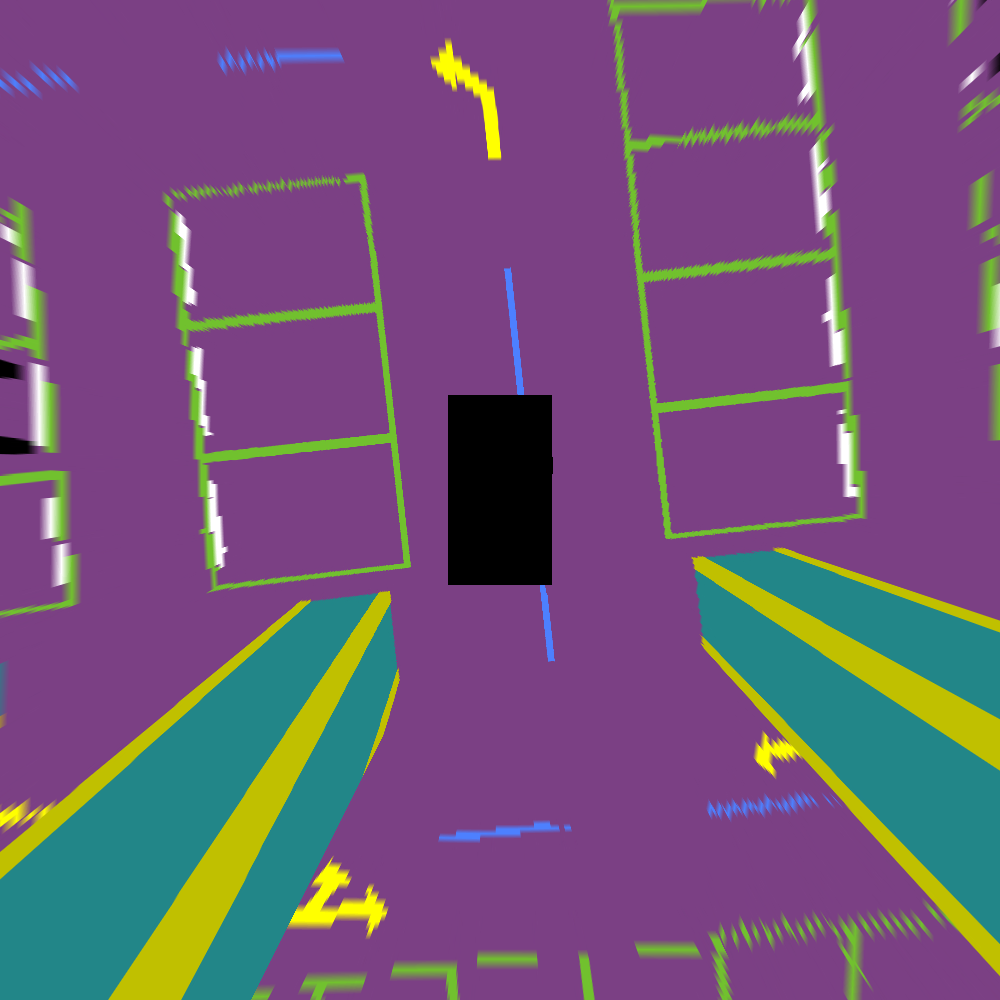}
    \includegraphics[height=2cm]{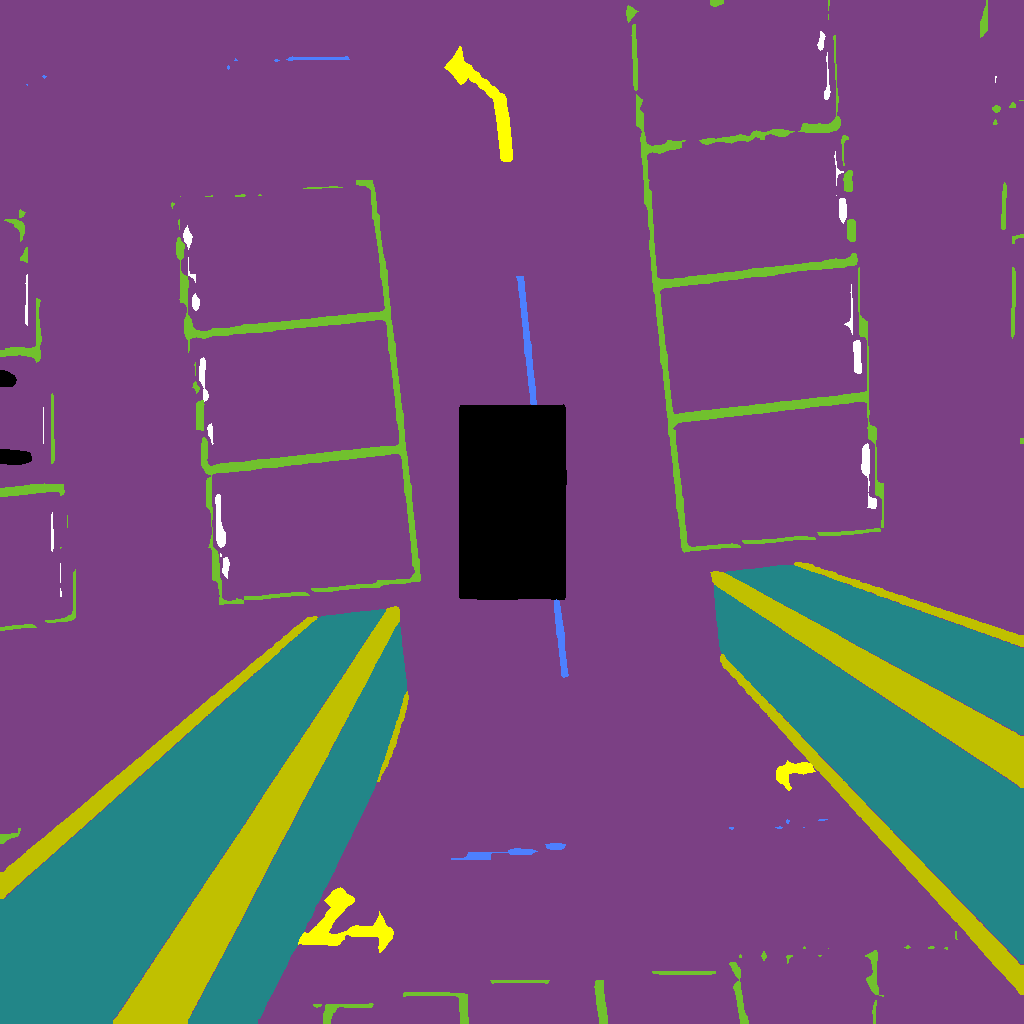}
    \includegraphics[height=2cm]{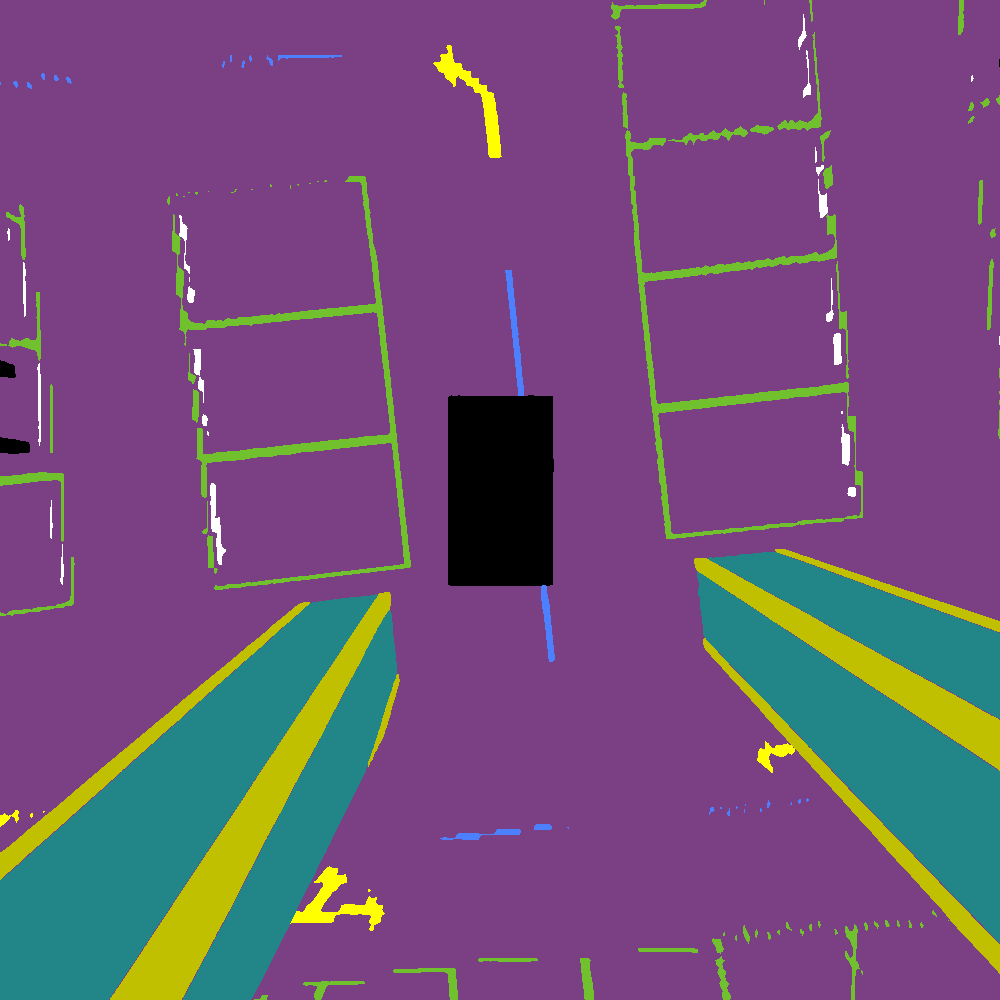}
    \caption{From the left side to right side, the first column is the original BEV image, the second column is the segmentation ground truth, the third column is the inference result from BiSeNet\cite{bisenet}, and the last column is the inference result from SFNet\cite{sfnet}.}
    \label{model_res}
\end{figure}

Semantic segmentation is a standard perception task, and numerous methods and networks have provided solutions to this issue. However, for autonomous driving usage, the focal point of this task focuses on the trade-off between accuracy and efficiency. BiSeNet\cite{bisenet} is a bilateral segmentation network including {\it Spatial Path} and {\it Context Path}, which achieves efficiency while ensuring accuracy. SFNet\cite{sfnet} proposes a flow alignment module to achieve feature fusion and achieves high mIoU on the Cityscape\cite{Cityscapes} test set. We assess the performance of both networks to test how well the algorithms can work on our dataset and the simultaneous usage practicability. The results can be seen in Tab. \ref{model_compare} and Fig. \ref{model_res}.

\begin{table}[t]
    \caption{Comparison of the MIOU and FPS between models.}
    \label{model_compare}
    \centering
    \begin{tabular}{c|ccc}
    \hline
        Model & MIOU & FPS \\
        \hline
        BiSeNet\cite{bisenet} & 0.8531 & 62.59 \\
        SFNet\cite{sfnet} & 0.9091 & 6.45 \\
        \hline
    \end{tabular}
    
\end{table}

\subsection{Visual-SLAM}
\begin{figure}
    \centering
    \subfigure[ORB-SLAM3\cite{orb3}]{
    \includegraphics[height=3cm]{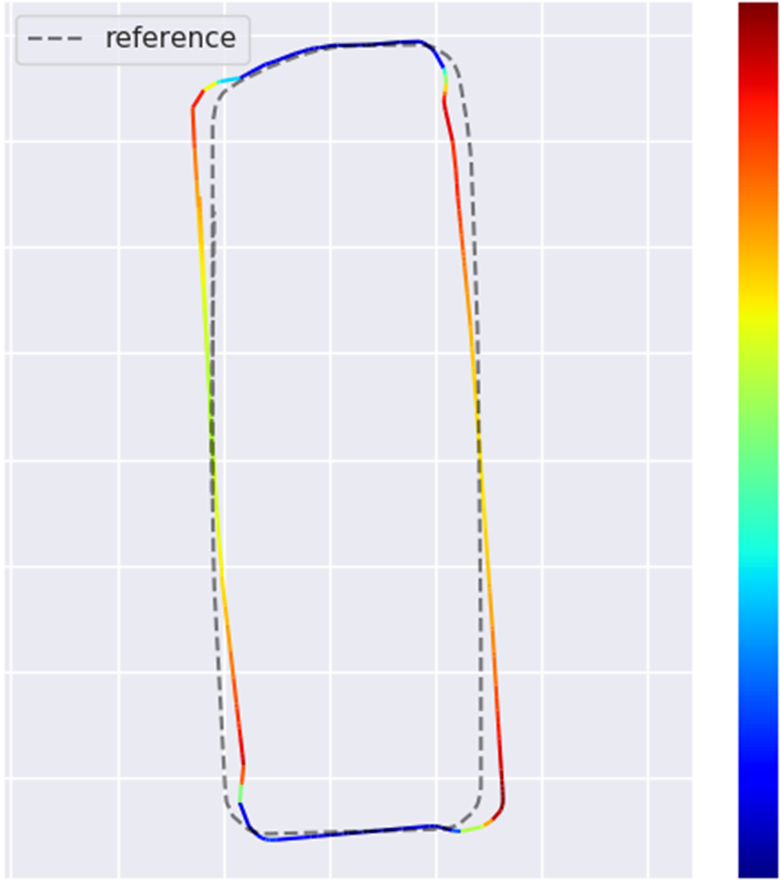}
    \includegraphics[height=3cm]{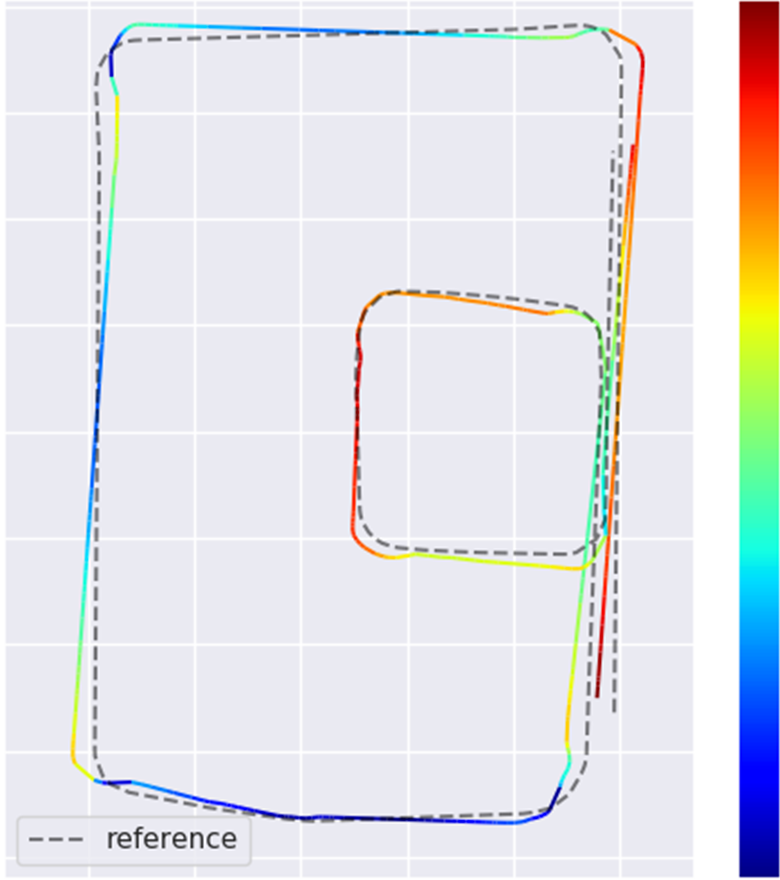}
    \includegraphics[height=3cm]{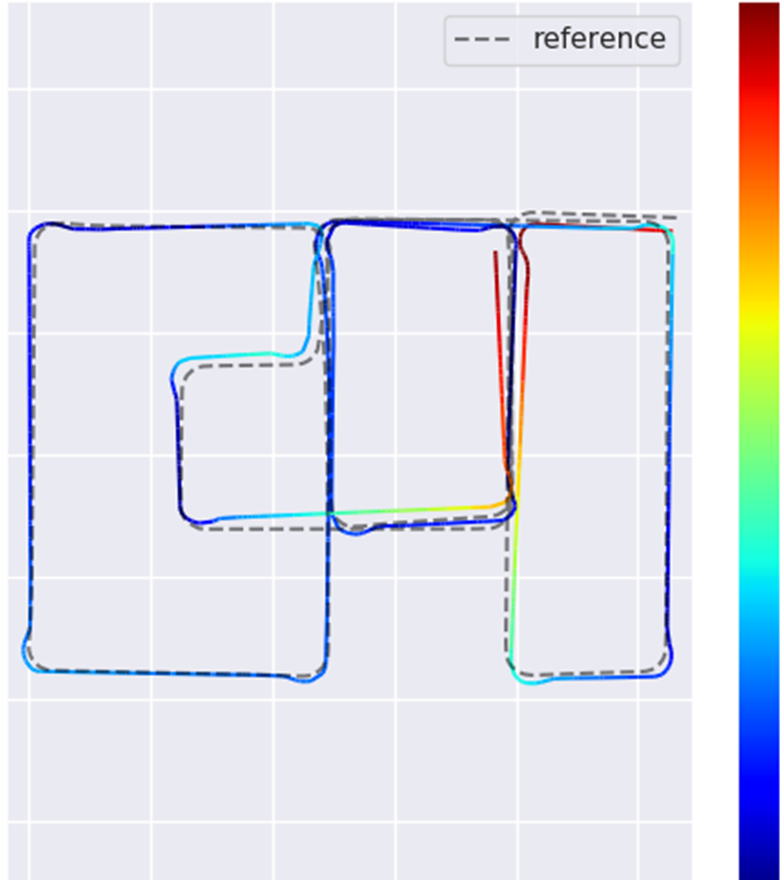}
    }
    \subfigure[VINS-Fusion\cite{vins-fusion}]{
    \includegraphics[height=3cm]{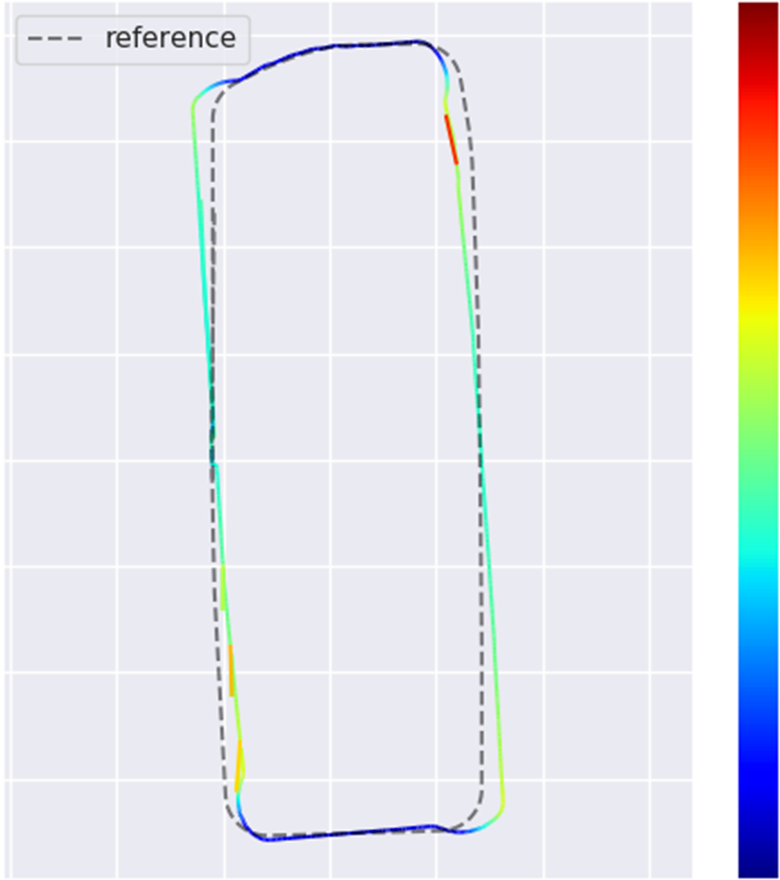}
    \includegraphics[height=3cm]{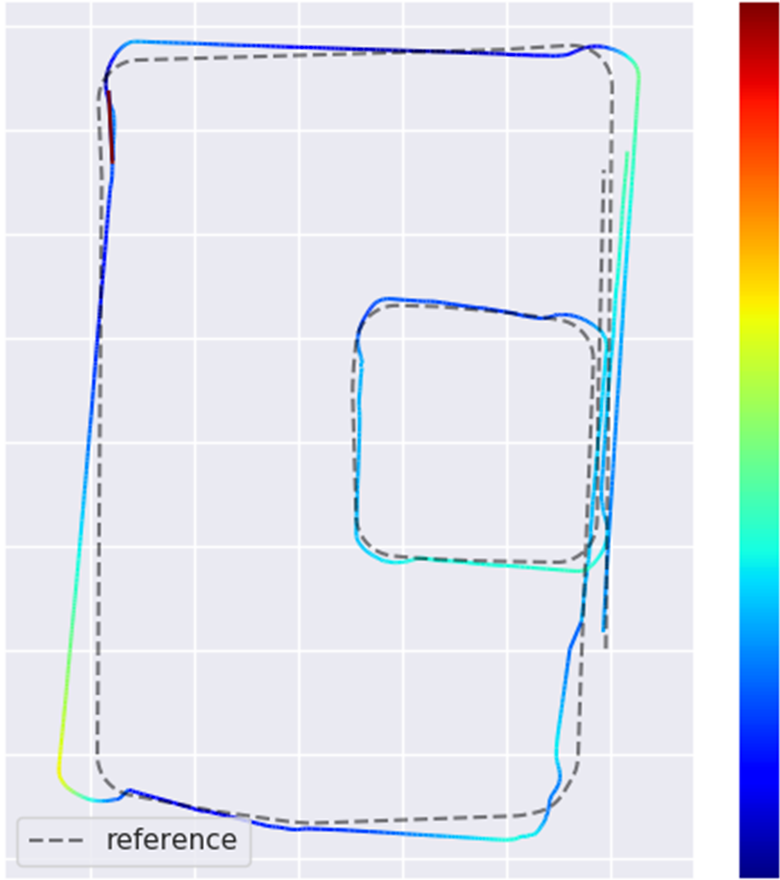}
    \includegraphics[height=3cm]{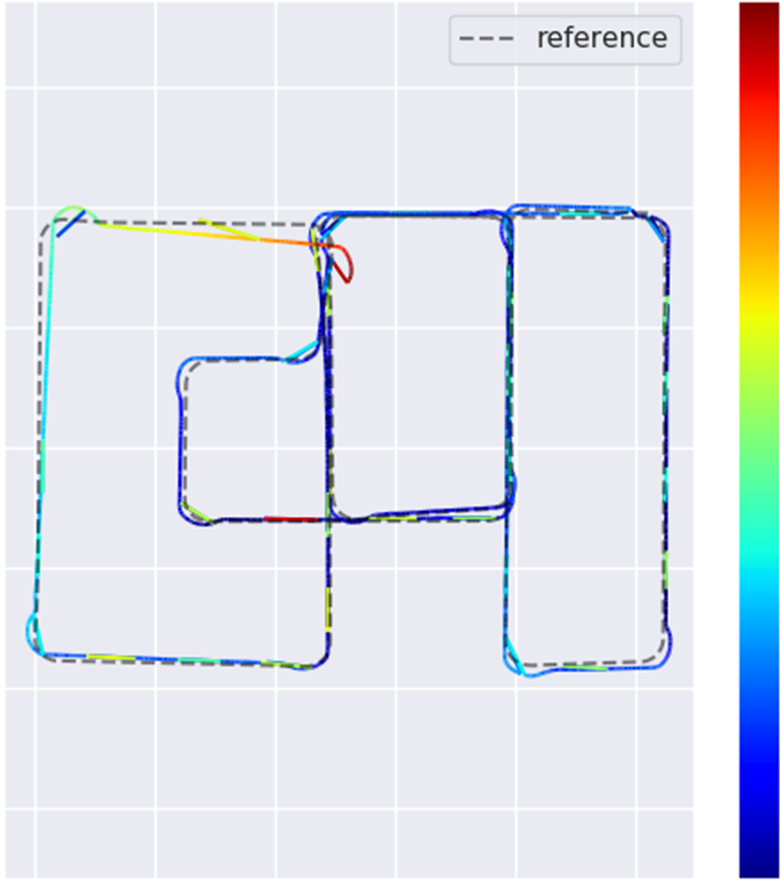}
    }
    \subfigure[LIO-SAM\cite{lio-sam}]{
    \includegraphics[height=3cm]{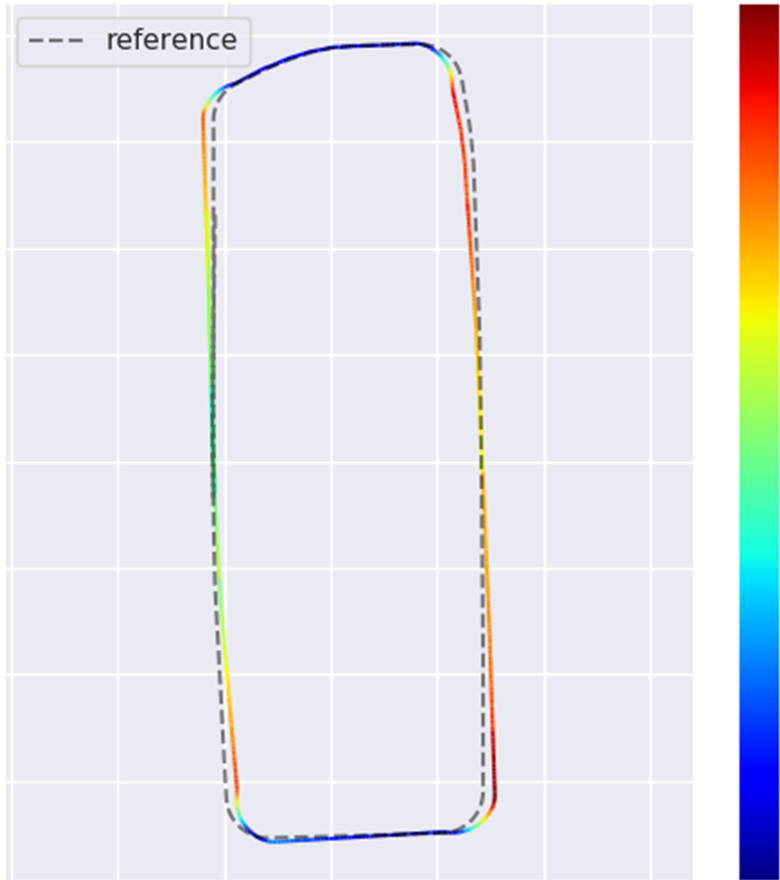}
    \includegraphics[height=3cm]{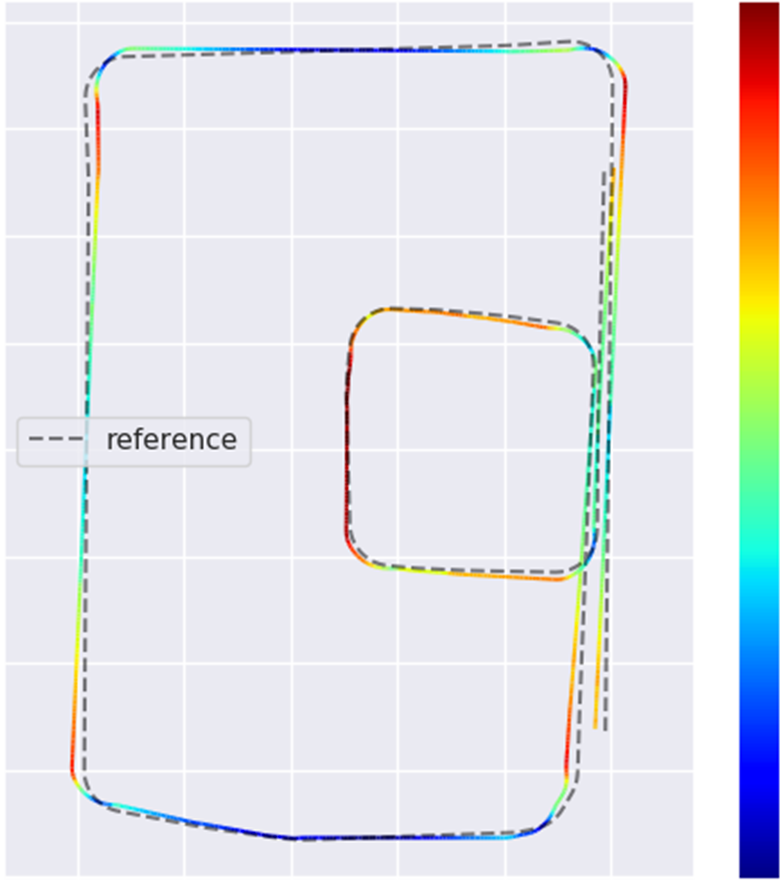}
    \includegraphics[height=3cm]{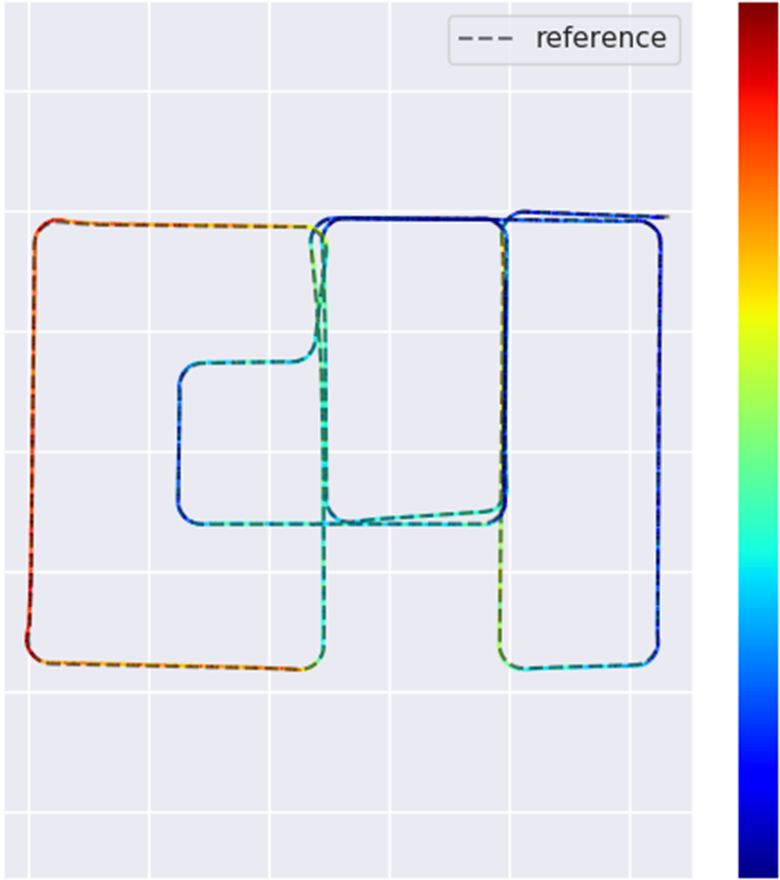}
    }
    \caption{Evaluation of the Visual SLAM and LiDAR based SLAM methods on three routes. The left column is evaluated on Loop A, the medial column evaluated on Loop B, and the right column evaluated on Loop C.}
    \label{eva_SLAM}
\end{figure}

The ORB-SLAM series is the open-source state-of-the-art Visual-SLAM solution, and ORB-SLAM3\cite{orb3} is the latest work of this series which has robust performance. VINS-Fusion\cite{vins-fusion} is another classical feature-based Visual-SLAM method. We benchmark their robustness and accuracy on our dataset on three routes. The experimental results are shown in Tab. \ref{ape_SLAM} and Fig. \ref{eva_SLAM}. We evaluate the absolute trajectory error (ATE) using EVO\cite{evo}. 


\begin{table}[]
\caption{Comparison of the APE error of the SLAM algorithms  on three routes.}
    \label{ape_SLAM}
    \centering
    \begin{tabular}{c|c|ccccc}
    \hline
        SLAM & Route & Max & Median & Min & RMSE & STD \\
        \hline
        \multirow{3}{*}{ORB3\cite{orb3}} & A & 2.70 & 1.86 & 0.17 & 1.78 & 0.77 \\
                                   & B & 3.12 & 1.88 & 0.23 & 1.96 & 0.78 \\
                                   & C & 10.25 & 3.14 & 1.14 & 4.78 & 2.57 \\
        \hline
        \multirow{3}{*}{VINS-F\cite{vins-fusion}} & A & 4.11 & 1.78 & 0.19 & 1.70 & 0.71 \\
                                     & B & 6.28 & 1.82 & 0.06 & 1.95 & 0.77 \\
                                     & C & 7.30 & 2.71 & 1.75 & 3.03 & 0.99 \\
        \hline
        \multirow{3}{*}{LIO-SAM\cite{lio-sam}} & A & 1.52 & 0.87 & 0.04 & 0.92 & 0.38 \\
                                 & B & 1.40 & 0.71 & 0.03 & 0.80 & 0.32 \\
                                 & C & 1.37 & 1.01 & 0.80 & 1.04 & 0.14 \\
        \hline
    \end{tabular}
    
\end{table}

\subsection{LiDAR-based SLAM}
While Visual-SLAM algorithms have more plentiful information and need fewer resources, LiDAR-based SLAM is a classical method that has reliable accuracy and robustness. LIO-SAM\cite{lio-sam} is a recent tightly coupled LiDAR SLAM method with IMU and GPS information. We evaluate the results of LIO-SAM\cite{lio-sam} and list the results in Fig. \ref{eva_SLAM} and Tab. \ref{ape_SLAM}.
\section{CONCLUSIONS}
\label{V}
We provide the SUPS dataset, a novel perception and SLAM benchmark that supports autonomous driving tasks in the underground parking scenario with complete sensors, e.g., surround fisheye cameras, forward pinhole cameras, depth camera, LiDAR, GNSS, and IMU. We evaluate the semantic segmentation, parking slot detection, and SLAM algorithms on the proposed dataset to verify its practicability. In addition to the SUPS dataset, we open-source the virtual scene and vehicle setups on the simulator for users making the specific adaptation.

Since we have provided a large FOV that can cover a complete parking space along the roadside in BEV sight, we have proposed a special description of the parking slot. One of our future directions is to develop object-level detection methods instead of using low-level corners and center lines to describe the parking slots. Object-level object detection could be more helpful than scattered structures in subsequent motion planning and decision-making procedures.



\addtolength{\textheight}{-7cm}   

\section*{Acknowledgment}
This work is supported by Shanghai Municipal Science and Technology Major Project (No.2018SHZDZX01), ZJ Lab, and Shanghai Center for Brain Science and Brain-Inspired Technology. 

\bibliographystyle{IEEEtran}
\bibliography{reference}

\end{document}